\documentclass[12pt]{cmuthesis}

\usepackage{times}
\usepackage{fullpage}
\usepackage{graphicx}
\usepackage{xcolor}
\usepackage{amsmath}
\usepackage[backend=biber]{biblatex}
\usepackage[acronym]{glossaries}
\usepackage[pageanchor=true,plainpages=false, pdfpagelabels, bookmarks,bookmarksnumbered,
]{hyperref}
\usepackage[nameinlink, capitalise]{cleveref}
\usepackage{subfigure}
\usepackage{multirow}

 \usepackage[T1]{fontenc}
\usepackage[extra,tone]{tipa}
\usepackage{amssymb}
\usepackage{algorithm}
\usepackage{algorithmic}
\usepackage{multicol}
\usepackage{lingmacros}
\usepackage{booktabs}
\usepackage{float}

\usepackage{tikz}
\usetikzlibrary{automata,arrows}

\usepackage[letterpaper,twoside,vscale=.8,hscale=.75,nomarginpar,hmarginratio=1:1]{geometry}

\newcommand{\olang}[1]{{\color{red} #1}}
\newcommand{\tlang}[1]{{\color{blue} #1}}
\newcommand{\ggglang}[1]{{\color{violet} #1}}
\newcommand{\gggsub}[2]{\{#1{\textgreater}#2\}}
\newcommand{\orth}[1]{$\langle$#1$\rangle$}

\newcommand{\langstack}[2]{
\begin{tabular}{c}
\olang{#1}\\
\tlang{`#2'}
\end{tabular}
}

\newcommand{\?}{\textipa{P}}

\makeatletter
\newcommand{\mapcolumn}[2]{
#1{#2} \@ifnextchar\bgroup{ & \maprec{#1}}{}
}
\newcommand{\maprec}[2]{#1{#2} \@ifnextchar\bgroup{ & \maprec{#1}}{}}

\makeglossaries

\newglossarystyle{lingabbr}{
  \setglossarystyle{long3colheader}%
  \setlength\glsdescwidth{.2\textwidth}%
  \setlength\glspagelistwidth{.7\textwidth}%
}

\setglossarystyle{lingabbr}

\newacronym[
  description={Referring to more than one of something}
]{pl}{\textsc{pl}}{Plural}

\newacronym[
  description={Referring to only one of something}
]{sg}{\textsc{sg}}{Singular}

\newacronym[
  description={(of a first-person plural) Including the listener}
]{incl}{\textsc{incl}}{Inclusive}

\newacronym[
  description={(of a first-person plural) Excluding the listener}
]{excl}{\textsc{excl}}{Exclusive}

\newacronym[
  description={Marks an event or action that occurs frequently}
]{hab}{\textsc{hab}}{Habitual}

\newacronym[
  description={A mood describing an event or action that is not known to have happened}
]{irr}{\textsc{irr}}{Irrealis}

\newacronym[
  description={Marks an event or action as a single, complete unit, without internal temporal structure}
]{cpl}{\textsc{cpl}}{Completive}

\newacronym[
  description={Indicates that an event or action does not happen}
]{neg}{\textsc{neg}}{Negative}

\newacronym[
  description={Marks a verb describing a state of being rather than an action}
]{est}{\textsc{stat}}{Stative}

\newacronym[
  description={Reduces the valency of a transitive verb, forming an intransitive verb}
]{detr}{\textsc{detr}}{Detransitive}

\newacronym[
  description={Tense describing current actions or events}
]{pres}{\textsc{pres}}{Present}

\newacronym[
  description={Reduces the valency of a verb, promoting the semantic role of patient to the subject position}
]{pass}{\textsc{pass}}{Passive}

\newacronym[
  description={A non-finite verb form that adopts some characteristics of adjectives}
]{part}{\textsc{ptcp}}{Participle}

\newacronym[
  description={Head of a dependent clause, such as ``that'' or ``which'' in English.}
]{comp}{\textsc{comp}}{Complementizer}

\newacronym[
  description={Derivation from a root noun into a verb.}
]{n>v}{\textsc{N>V}}{Noun to Verb}

\newacronym[
  description={Derivation from a root noun into an adjective.}
]{n>adj}{\textsc{N>Adj}}{Noun to \mbox{Adjective}}

\newacronym[
  description={Derivation from a root verb into a noun.}
]{v>n}{\textsc{V>N}}{Verb to Noun}

\newacronym[
  description={Derivation from a root adjective into an noun.}
]{adj>n}{\textsc{Adj>N}}{Adjective to Noun}

\newacronym[
  description={A locative case carrying the meaning ``in'' or ``at'' (used in Finnish)}
]{ine}{\textsc{ine}}{Inessive}

\newacronym[
  description={Expresses that the subject makes the object adopt a certain quality or end state (used in Finnish)}
]{fact}{\textsc{fact}}{Factitive}

\begin {document} 
\frontmatter

\pagestyle{empty}

\def\maintitle{Morphologically-Informed Tokenizers for Languages with Non-Concatenative Morphology: \\ {\Large A case study of Yolox\'ochtil Mixtec ASR}}

\title{ %
{\Large\bf \maintitle}}
\author{Chris Crawford}
\date{December 2025}
\Year{2025}
\trnumber{CMU-CS-25-142}

\committee{
David Mortensen, Chair \\
Shinji Watanabe
}

\department{Computer Science Department}
\degree{Masters degree in Computer Science}

\support{}
\disclaimer{}


\maketitle

\begin{dedication}
For all those who work and dedicate themselves to ensure everyone and every culture has access to what they need to participate and thrive in a technology-driven world.
\end{dedication}

\pagestyle{plain} %

\begin{abstract}
This paper investigates the impact of using morphologically-informed tokenizers to aid and streamline the interlinear gloss annotation of an audio corpus of Yolox\'ochitl Mixtec (YM) using a combination of ASR and text-based sequence-to-sequence tools, with the goal of improving efficiency while reducing the workload of a human annotator. We present two novel tokenization schemes that separate words in a nonlinear manner, preserving information about tonal morphology as much as possible. One of these approaches, a Segment and Melody tokenizer, simply extracts the tones without predicting segmentation. The other, a Sequence of Processes tokenizer, predicts segmentation for the words, which could allow an end-to-end ASR system to produce segmented and unsegmented transcriptions in a single pass. We find that these novel tokenizers are competitive with BPE and Unigram models, and the Segment-and-Melody model outperforms traditional tokenizers in terms of word error rate but does not reach the same character error rate. In addition, we analyze tokenizers on morphological and information-theoretic metrics to find predictive correlations with downstream performance. Our results suggest that nonlinear tokenizers designed specifically for the non-concatenative morphology of a language are competitive with conventional BPE and Unigram models for ASR. Further research will be necessary to determine the applicability of these tokenizers in downstream processing tasks.
\end{abstract}

\begin{acknowledgments}

Thanks to David Mortensen, for being my inspiration and my research advisor for both undergraduate and graduate school. 

Thanks to Jonathan D. Amith and Rey Castillo Garc\'ia for allowing us to work with their data. Amith and Garc\'ia provided the dataset for Yolox\'ochitl Mixtec used in this project. It is a continuation of their YM audio corpus published in 2021 \cite{shi2021leveragingendtoendasrendangered}, along with an unpublished YM dictionary with 2922 lexical entries as of August 2025 \cite{amithdictionary}, with updates that are still in progress at the time of writing. Our team assisted with the curation of ambiguous sentences, data cleanup and normalization, and tools to assist with the G3 and gloss annotations.

Thanks to Kalvin Chang for first referring me to the wav2gloss project. %

Thanks to Charlotte Li for helping with the project over the summer, particularly with finetuning the ByT5 model for segmentation and curating the samples for the evaluation dataset.

Thanks to Lori Levin for saving my Master's program and keeping my research on track as I navigated logistical difficulties.

Thanks to Jiatong Shi, Taiqi He, Lindia Tjuatja, Morris Alper, and the other students I've had the honor of working with over the course of this project, whose current and prior research was pivotal in producing this thesis.

Thanks to Michael Hilton for recommending I take the Fifth Year Master's program.

Thanks to Tracy Farbacher for helping me through the logistics of the Fifth Year Program.

Thanks to Angy Malloy for keeping me on through the fall term, and for being flexible in understanding that data doesn't always arrive within hard deadlines.

Thanks to my parents for believing in me and being there for me to lean on when I need them most.

\end{acknowledgments}

\tableofcontents
\listoffigures
\listoftables

\mainmatter

\chapter{Introduction}

In this paper, we will investigate the impact of using tailored, morphologically-informed tokenizers on ASR performance, as well as the extent to which an ASR model can predict segmentations and morphological processes, for a low-resource language that exhibits a large amount of non-concatenative morphology. We predict that by separating the tones and tonal morphology from the root segments, our models will be better able to parse the morphology of Yolox\'ochitl Mixtec and, in turn, achieve better performance in ASR and downstream tasks.

We are exploring two research questions:
\begin{itemize}
    \item How does the use of morphologically-informed tokenizers affect ASR performance?
    \item How well can an ASR model be trained to predict morphological segmentations?
\end{itemize}

This project seeks to improve models by developing customizable, language-specific tokenizers to complete the interlinear gloss annotation of an audio corpus of Yolox\'ochitl Mixtec using a combination of ASR and text-based sequence-to-sequence tools.

\section{Tokenization}
\label{intro:tokenization}

In NLP and speech processing, \textbf{tokenization} is the process of converting a text string into a sequence of integers that a language model can process natively. Each of these integers and their associated strings is called a \textbf{token}. Crucially, tokenization should be done without losing information, so that the original text can be reconstructed unambiguously from the integer sequence.  There are many approaches to tokenization, including naive methods such as assigning each character to be a token (character-level tokenization) or splitting on whitespace and assigning each word to be a token (word-level tokenization). However, each of these approaches has a trade-off: word-level tokenizers produce more compact and meaningful representations but aren't able to recognize words outside of their vocabulary, while character-level tokenizers can represent any input string, though they also produce long sequences of tokens that are too abstract for models to meaningfully represent.

In order to strike a balance and develop a tokenizer that is capable of both encoding texts with unseen vocabulary and producing a relatively compact sequence of meaningful tokens, we should consider \textbf{subword tokenization}, in other words a tokenizer that may produce multiple tokens per word, but has longer and more meaningful tokens than simply using characters. Linguistic principles suggest that there is merit to this approach, as many words can be broken down into smaller units that still carry a discernible meaning. For example, the English word ``\olang{unbelievable}'' can be broken down as ``\olang{un}-\olang{believ(e)}-\olang{able}'', each of which carries some meaning: ``\olang{un}-'' means ``not'', ``-\olang{able}'' means ``can be done'' or ``inclined to'', and the root word ``\olang{believe}'' has a transparent meaning. Linguists refer to these meaningful subword units as \textbf{morphemes}. Morphemes can perform many different functions within a language:
the suffix ``\olang{-s}'' in the English word ``\olang{dogs}'' is an example of an \textbf{inflectional} morpheme---one which indicates grammatical information---whereas the affixes ``\olang{un-}'' and ``\olang{-able}'' in the English word ``\olang{unbelievable}'' are examples of \textbf{derivational} morphemes--ones which are used to build new words out of existing words.

The linguistic field of \textbf{morphology}---the study of words and their structure---is very insightful when designing a subword tokenization method. Ideally, tokenizers should leverage \textbf{compositionality}, building a representation of a word from representations of it's constituent parts in a way that reflects how the meaning of a word can be understood from the meanings of its constituent morphemes. Understanding morphology gives us a way to evaluate the efficacy of a subword tokenizer, with better tokenizers correlating well with a language's morphology.

One popular method of subword tokenization is byte-pair encoding, or BPE.
The byte-pair encoding (BPE) algorithm starts with a set of character-level tokens and a training corpus. At each step, it finds the most frequent pair of consecutive tokens within each word of the corpus and merges them into a single token. For instance, if the training corpus consists of the sentence ``\olang{These foxes are in boxes}'', then the first merge would be the characters \texttt{e} and \texttt{s} into the token \texttt{es}. The next merge would be \texttt{o} and \texttt{x} into the token \texttt{ox}, followed by merging the tokens \texttt{ox} and \texttt{es} into \texttt{oxes}. After this sequence of merges, there are no more pairs of tokens that occur more than once, so stopping the algorithm here would yield the following tokenization:

\[\texttt{\textunderscore T h es e \textunderscore f oxes \textunderscore a r e \textunderscore i n \textunderscore b oxes}\]

The merging step is repeated until the vocabulary size, a hyperparameter set by the user, is reached, resulting in a set of tokens that (1) captures common subword strings, (2) can encode any piece of text within the same character set as the training corpus, and (3) reduces the expected length of an encoded sequence by merging the most frequent character $n$-grams, up to and including full words, into single tokens \cite{sennrich-etal-2016-neural}.

While BPE is not an optimal tokenizer in all tasks, it exhibits competitive performance in practice for languages that exhibit concatenative morphology \cite{amrhein2021suitablesubwordsegmentationstrategies,bostrom2020byte}. This competitive performance, combined with the relative ease of computation compared to most alternative tokenizers, contribute to its overall popularity.

\section{Glossing}

Interlinear glossed text (IGT) is a type of linguistic annotation that informs readers about the mapping from form to function within a language, typically following a set of conventions, such as the Leipzig glossing rules \cite{comrie2008leipzig}, to provide a common format for linguists and language teachers to communicate about the languages they study. Using IGT, a linguist can describe the \textbf{object language}--the language being studied--in terms of the \textbf{metalanguage}--the language used to write the paper or annotate the resource--without losing or obscuring information. Additionally, glosses can be written to different levels of granularity. Sentences~(\ref{sent:gloss-example-basic}) and (\ref{sent:gloss-example-deep}) below provide an example gloss for a phrase in {Finnish} from the Parallel Universal Dependencies dataset \cite{PUD-zeman-EtAl:2017:K17-3}, with the object language annotated in red, the metalanguage gloss in black, and the idiomatic free translation in blue. The sentence itself is the same in both, but the former provides a basic segmentation of basic compounds and inflection--grammatical information explaining the relationships among words in a sentence--while the latter provides a deeper segmentation of the derivational morphology--the internal structure of each word and how it is built up from other words.

\eenumsentence{
\item
\label{sent:gloss-example-basic}
\shortex{2}{
\mapcolumn{\olang}{Toteutettavuus-tutkimu-ksessa}{arvioi-daan}%
}{
feasibility-study-\acrshort{ine}.\acrshort{sg} & estimate-\acrshort{pass}
}{
\tlang{`The feasibility study estimates ...'}
\label{sent:gloss-example}
}
\item
\label{sent:gloss-example-deep}
\shortex{2}{
\mapcolumn{\olang}{Tote-ut-ettav-uus+tutki-mu-ksessa}{arvi-oi-daan}%
}{
true-\acrshort{fact}-\acrshort{pass}.\acrshort{pres}.\acrshort{part}-\acrshort{adj>n}+research-\acrshort{v>n}-\acrshort{ine}.\acrshort{sg} & estimation-\acrshort{n>v}-\acrshort{pass}
}{
\tlang{`The feasibility study estimates ...'}
}
}

The example in (\ref{sent:gloss-example-basic}) is sufficient for teaching purposes, providing a breakdown of root forms and inflectional morphology for a language learner to get a sense of the important structures in the language. The example in (\ref{sent:gloss-example-deep}), on the other hand, peels back more layers in the derivation and etymology of these forms, which would be useful to a linguist studying morphology in developing a deeper understanding of word formation in Finnish. Linguists might also use a gloss like (\ref{sent:gloss-example-basic}) if word formation and morphological structure is not their primary focus: a paper on syntax and sentence structure need not peel back the derivational morphology of the compound for `feasibility study', as that has no impact on the word's role within the sentence.

Interlinear glosses are interesting to us for this task because they provide a representation of the morphology of some object language in a manner that does not require understanding the object language itself. The segmentation and gloss together provide information on how subword units combine to create meaning in the object language, which can in turn be used to inform the development for tokenizers in that language.

\chapter{Background}

\section{The wav2gloss Task}

Glossing has been around for centuries, with IGT in particular tracing its roots to the early to mid 19th century \cite{humboldt1836verschiedenheit}. It remains an invaluable tool for linguists, though producing IGT annotations is very labor-intensive. Automating the process of producing IGT annotations would greatly speed up the process of compiling IGT corpora and, in turn, improve access to valuable and insightful data, particularly in low-resource languages.

The wav2gloss task is to predict IGT directly from speech audio.
A major application of this task is the documentation and preservation of endangered languages. The process of documentation involves a combination of recording audio of the language as it is spoken, transcribing the audio into text, and annotating the semantic and morphological content of each utterance into the metalanguage. However, while collecting audio corpora is relatively easy, transcribing and annotating the recorded data is very time consuming: each minute of spoken data can require an hour of work by a trained linguist and a native-speaker consultant just to produce the object language transcription \cite{do:halshs-00980431}. This bottleneck severely limits the scope of fully annotated corpora to a small subset of the available audio. Using ASR to produce a draft transcription has long been recommended, as it reduces the time a human annotator needs to produce a final transcription: revising an ASR-generated transcription for mistakes takes less time than manual transcription of the whole corpus. Previous experiments by Amith et al. have shown that this approach is effective, with a reduction in effort of 75\% by human annotators when utilizing an ASR system capable of outputting a transcription with a character error rate (CER) below 10 percent, which is considered to approach the minimum amount of effort to proofread transcribed speech in an endangered language \cite{amith2021end}.

Given the positive results thus far with machine-assisted transcription, it is also worth exploring the possibility of outputting other types of annotations, such as segmentations, glosses, and even free translations, which we expect will further accelerate the process of annotating spoken-language corpora. Results by He et al. on this task suggest that a language-specific model for segmentation may be feasible, although this problem remains unsolved. Their baselines for end-to-end systems to output the underlying forms and glosses exhibit considerably higher error rates than the sequence-to-sequence models working directly from ground truth text transcriptions \cite{he2024wav2glossgeneratinginterlinearglossed}.

Leveraging ASR to predict morphological segmentations provides a path forward. If a model can learn to perform morphological segmentations with high accuracy, then it can be used both for producing a meaningful breakdown of the morphemes, which can be passed to a downstream glossing model, as well as the simpler unsegmented transcription. We can expect that a glossing model will be better able to directly predict the glosses and meanings from a segmented transcription than from an unsegmented transcription, as shown by results from Ginn et al., 2024 \cite{ginn2024glosslmmultilingualpretraininglowresource}.

\section{Generalized Glossing Guidelines (G3)}

The Generalized Glossing Guidelines (G3) is a novel IGT format designed to be both unambiguous and adaptable in its representation of non-concatenative morphology, and to be easily readable to both humans and machines \cite{mortensen2023generalized}. This format maintains common conventions for handling concatenative morphology from other glossing formats, such as separating affixes with a hyphen \orth{-} and clitics\footnote{Clitics are a type of morpheme that are realized as affixes, but syntactically behave like independent words.} with an equals sign \orth{=}, while also introducing an arrow notation in curly braces that can represent various kinds of non-concatenative processes, as illustrated in Figure~\ref{fig:g3-examples}. The format of this notation is \texttt{\{A>B\}}, where \texttt{A} and \texttt{B} are potentially empty strings. The string to the left of the arrow, \texttt{A}, represents the original text from the morphological representation, and the string to the right of the arrow, \texttt{B}, represents the replacement as a result of the morphological process. This allows us to reconstruct both the morphemic representation, eg. Yolox\'ochitl Mixtec \orth{\olang{sa3ta4}} meaning \tlang{`to buy'} from (\ref{sent:g3-mixtec}), as well as the underlying representation, eg. \orth{\ggglang{sa\{4\}ta\{2\}4}} meaning \tlang{`habitually buy'}.
\begin{figure}
\begin{multicols}{2}
\enumsentence{
\shortex{1}{
\mapcolumn{\ggglang}{Br\gggsub{u}{\"u}der}
}
{brother\{\acrshort{pl}\}}
{\tlang{`brothers'}}
\textit{Apophony (German)}
}

\enumsentence{
\shortex{1}{
\mapcolumn{\ggglang}{s\gggsub{}{um}ulat}
}
{writing\{\acrshort{cpl}\}}
{\tlang{`to write'}}
\textit{Infixation (Tagalog)}
}

\enumsentence{
\shortex{1}{
\mapcolumn{\ggglang}{\gggsub{}{ma}mahuta}
}
{sleep\{\acrshort{pl}\}}
{\tlang{`to sleep'}}
\textit{Reduplication (Motu)}
}

\enumsentence{
\label{sent:g3-arabic}
\shortex{1}{
\mapcolumn{\ggglang}{k\gggsub{i}{u}t\gggsub{a:}{u}b}
}
{book\{\acrshort{pl};1,2\}}
{\tlang{`books'}}
\textit{Transfixation (Arabic)}
}

\enumsentence{
\label{sent:g3-mixtec}
\shortex{1}{
\mapcolumn{\ggglang}{sa\gggsub{3}{4}ta\gggsub{}{2}4}
}
{buy\{\acrshort{hab};1,2\}}
{\tlang{`habitually buy'}}
\textit{Tonal overwriting (Mixtec)}
}

\enumsentence{
\shortex{1}{
\mapcolumn{\ggglang}{\gggsub{xi}{ku}3xi3}
}
{eat\{\acrshort{irr}\}}
{\tlang{`to eat'}}
\textit{Segmental overwriting (Mixtec)}
}
\end{multicols}
    \caption{Examples of the G3 format in various languages, showcasing different non-concatenative processes it can represent.\cite{mortensen2023generalized}}
    \label{fig:g3-examples}
\end{figure}

Formally, G3 defines four transcription tiers: 
\begin{itemize}
    \item \texttt{lx}: A lexical or morphemic representation, explicitly showing each morpheme in its underlying, canonical form.
    \item \texttt{sr}: A surface representation, which reflects the phonemic form after both morphological and phonological rules have been applied.
    \item \texttt{gl}: A metalanguage gloss of each token and process in the \texttt{lx} and \texttt{sr} strings.
    \item \texttt{tr}: An idiomatic free translation of the glossed discourse segment into the metalanguage.
\end{itemize}

A key feature of the G3 tiers is the association between each annotated process and its associated gloss. A morphological process that covers multiple annotated rewrite spans is indexed with the corresponding spans, as seen in sentences~\ref{sent:g3-arabic} and \ref{sent:g3-mixtec}. 

The lexical representation, \texttt{lx}, is of particular interest to this project. Its robustness allows us to build models that generate other tiers from \texttt{lx}, and as such we will consider this form as the gold standard for our segmentation task.

\section{Tokenization}
\label{bg:tokenization}

As introduced in section \ref{intro:tokenization}, byte-pair encoding (BPE) is a widely-used tokenization method that breaks text into single-byte tokens\footnote{Assuming byte-wise BPE. Character-wise BPE, another implementation, doesn't split apart characters that are encoded with two or more bytes.}, then repeatedly merges common pairs of tokens until some set vocabulary size has been reached. This approach is effective at modeling concatenative morphology, as the tokenizer naturally picks up on frequently occurring substrings in a corpus, which tend to correspond with common prefixes, suffixes, and roots in the language. Although BPE tokens do not always correspond with morpheme boundaries, as exemplified by the BPE tokenization of the Finnish word in Figure~\ref{fig:bpe-g3-fi-ym}, the general correspondence of each token to a meaningful unit in the language tends to improve language model performance. Models built with other tokenizers, such as Unigram and WordPiece, sometimes outperform those built with BPE, and this performance improvement correlates with the tokenizers predicting morphology more accurately than BPE. In 2020, Bostrom and Durrett showed that using Unigram tokenization for pretraining language models in English and Japanese results in better performance for downstream tasks than using BPE across all tasks. They also showed that the subword token boundaries predicted by the Unigram tokenizer exhibited better correspondence with morphological segmentations than those predicted by the BPE tokenizer in both languages \cite{bostrom2020byte}.

Toraman et al. compared different tokenization methods for Turkish, including BPE, morphological segmentation, and WordPiece--a similar algorithm to BPE that optimizes a different heuristic to identify which tokens to merge \cite{toraman2023impact}. They found BPE and WordPiece to be the best at modeling Turkish, with WordPiece scoring better than BPE in five out of six tasks. Interestingly, the higher-level morphological tokenization they implemented did not perform as well as BPE or WordPiece, although it was competitive. The authors remarked that the morphological tokenizer had two main issues that likely hindered performance. First, the tokenizer was less robust than BPE and WordPiece at parsing out-of-vocabulary terms, being limited to word stems already in its dictionary and unable to parse new roots.
Second, they faced issues with the morphological analysis tool they used, Zemberek \cite{akin2007zemberek}, 
which was not parsing certain words completely, such as treating \orth{\olang{\.Istanbullular}} as a single token despite comprising derivational and inflectional affixes as shown in (\ref{sent:istanbul}). Consequently, the intuition that tokenizing based on a robust morphological model produces better-performing language models is still theoretically sound, so long as the tokenizer is robust at recognizing both internal morphological structure and out-of-vocabulary items.

\enumsentence{
\shortex{3}{\mapcolumn{\ggglang}{\.Istanbul}{-lu}{-lar}}{Istanbul & -person.from & -\acrshort{pl}}{\tlang{`The people of Istanbul'}}
\label{sent:istanbul}
}

Another hurdle for our project is that most work in tokenization considers only concatenative morphology, with only a handful of papers considering non-concatenative morphology. 
A 2021 paper by Amrhein and Sennrich compared three BPE models and one character-level model in their ability to model non-concatenative morphology in translation tasks, particularly infixation, vowel harmony, and reduplication \cite{amrhein2021suitablesubwordsegmentationstrategies}. In order to ensure high translation reliability, they augmented a corpus of ~4.6M sentences of English-German parallel data to include artificially generated non-concatenative morphemes. They found that the BPE model with 500 merge operations was more accurate in capturing vowel harmony and reduplication than the BPE models with 32k merge operations, and also that the character-level model performed similarly well to the BPE-500 model. These results indicate that tokenization with smaller units provides a benefit to modeling non-concatenative morphology.
Additionally, they tested an abstract representation in which the non-concatenative morphemes were replaced with separate words, such as \texttt{@INFIX{\textunderscore}4@} following the infixed word rather than using the infix itself as in the surface representation. Running the basic BPE-32k model over this tier of the dataset yielded competitive results across the board, with the model trained on the abstract representation outperforming the best models trained on the surface representation for infixation and vowel harmony in the target language. This result directly informed our approach, as it presented a way to run a nonlinear transformation to separate non-concatenative morphemes into separate abstract tokens.

A 2025 paper by Asgari et al. introduced MorphBPE, a novel extension to BPE that only performs merges that do not cross morpheme boundaries \cite{asgari2025morphbpemorphoawaretokenizerbridging}. The F1 scores for consistency between morphemes and predicted segments were higher for MorphBPE than standard BPE in all languages, with the extent of improvement correlating with the morphological complexity of the language. In particular, their experiments on Arabic, a language with non-concatenative templatic morphology that the authors labeled as having a ``highly complex'' morphology, this score increased from nearly 0.00\footnote{Scores were computed through $k$-means clustering and bootstrapping, for which they calculated a mean precision of $0.00 \pm 0.00$ and a mean recall of $0.08 \pm 0.03$ \cite{asgari2025morphbpemorphoawaretokenizerbridging}. The F1 score presented in the paper is exactly zero up to $\pm0.005$.} to 0.66. These results are somewhat expected, as the MorphBPE tokenizer is specifically designed to match morphology more closely. In addition, tracking cross-entropy loss for models trained on MorphBPE and standard BPE showed consistent improvement in models trained on MorphBPE over those trained on standard BPE, with marked improvement for Russian and Arabic, both of which have a moderately to highly complex morphology and exhibited low morphological consistency with standard BPE tokenizers. For Hungarian, an agglutinative language that was also labeled as having a highly complex morphology, model performance did not improve as much by using MorphBPE. However, Hungarian also had the highest morphological consistency score for standard BPE out of all the languages tested, suggesting that standard BPE was able to pick up on much of the important morphology in Hungarian.

One reason standard BPE may struggle to accurately model languages with non-concatenative morphology is that the algorithm produces tokens that do not reflect the underlying morphology. For example, if the YM word \orth{\olang{ta'14bi4}} is split into the tokens \texttt{ta'1 {\#\#}4bi4}, the connection to the lemma, \orth{\olang{ta'3bi4}}, is lost, while unrelated words that coincidentally begin with \texttt{ta'1} or contain \texttt{4bi4} will influence the model. The actual segmentation according to the BPE tokenizer from \cite{shi2021leveragingendtoendasrendangered} is given in Figure~\ref{fig:bpe-g3-fi-ym}. 

\enumsentence{
\label{sent:tabi-tokenizer}
\shortex{1}{\ggglang{ta'\{3>14\}bi4}}{break\{\acrshort{neg}\}}{\tlang{`To not break (something)'}}
}

\begin{figure}
    \centering
    
    \newcommand{\hz}{\vphantom{B}}
    \begin{tabular}{cc}
       \large \colorbox{cyan}{\hz\hspace{-0.1667em}T\hspace{-0.1667em}}\colorbox{yellow}{\hz\hspace{-0.1667em}ot\hspace{-0.1667em}}\colorbox{orange}{\hz\hspace{-0.1667em}e\hspace{-0.1667em}}\colorbox{lime}{\hz\hspace{-0.1667em}ut\hspace{-0.1667em}}\colorbox{pink}{\hz\hspace{-0.1667em}etta\hspace{-0.1667em}}\colorbox{cyan}{\hz\hspace{-0.1667em}v\hspace{-0.1667em}}\colorbox{yellow}{\hz\hspace{-0.1667em}uu\hspace{-0.1667em}}\colorbox{orange}{\hz\hspace{-0.1667em}st\hspace{-0.1667em}}\colorbox{lime}{\hz\hspace{-0.1667em}ut\hspace{-0.1667em}}\colorbox{pink}{\hz\hspace{-0.1667em}kim\hspace{-0.1667em}}\colorbox{cyan}{\hz\hspace{-0.1667em}uksessa\hspace{-0.1667em}} & \large \colorbox{cyan}{\hz\hspace{-0.1667em}ta\hspace{-0.1667em}}\colorbox{yellow}{\hz\hspace{-0.1667em}'1\hspace{-0.1667em}}\colorbox{orange}{\hz\hspace{-0.1667em}4\hspace{-0.1667em}}\colorbox{lime}{\hz\hspace{-0.1667em}bi\hspace{-0.1667em}}\colorbox{pink}{\hz\hspace{-0.1667em}4\hspace{-0.1667em}} \\
       \ggglang{\large Tote- ut- ettav- uus- tutki -mu -ksessa}  & \ggglang{\large ta'\gggsub{3}{14}bi4}
    \end{tabular}
    \caption{Comparison of BPE tokenization and formal segmentation for Finnish (left) and Yolox\'ochitl Mixtec (right). Each BPE token is represented by a colored box, and segmentation is presented in G3. Finnish example is glossed in Sentence~(\ref{sent:gloss-example-deep}), and YM example is glossed in Sentence~(\ref{sent:tabi-tokenizer})}
    \label{fig:bpe-g3-fi-ym}
\end{figure}

The BPE tokenizer from \cite{shi2021leveragingendtoendasrendangered} generates tokens that reflect the segments of YM words fairly well, but tends to group tones with neighboring characters that do not carry individual meaning, such as a preceding glottal stop \orth{'} or a following clitic marker \orth{=}. Despite the predicted analysis of the clitic marker to correspond with the following enclitic, since clitics in YM are exclusively suffixing, the more frequent pairing of the literal \texttt{=} character and a preceding tone number results in the BPE algorithm picking up on this correspondence and merging the tokens before picking up on the correspondence of the equals sign and the common enclitics.

We believe that this inadequacy of standard BPE-style tokenization in representing non-concatenative morphology represents a gap. Expanding on prior research, we consider tokenizers that apply morphologically-informed, nonlinear transformations to improve processing in languages like YM that exhibit non-concatenative morphology.

\section{Finite-State Transducers}

Having explored the state-of-the-art in linear tokenization methods, this section shifts to exploring a formalism that can model nonlinear, rule-based text processing. Finite-State Transducers (FSTs) model rational relations between regular languages, which is sufficient expressive power to represent morphological segmentation rules, including non-concatenative rules. As such, we consider FSTs as one possible avenue to 

Finite-State Automata (FSAs) are abstract machines with a finite amount of states, that transition from state to another as they read inputs. FSAs have long been studied in computer science as an abstraction for certain types of computation, as well as their provable equivalence to regular languages.
Finite-State Transducers (FSTs), a type of FSA, map input strings to output strings by following the state transitions based on the input \cite{mohri1996some}. Formally, we can define an FST to be a 7-tuple $\mathcal{T}=\langle Q, q_0, F, \Sigma_a,\Sigma_b, \delta, \sigma\rangle$, where
\begin{itemize}
    \item $Q$ is the set of states,
    \item $q_0\in Q$ is the unique\footnotemark initial state,
    \item $F \subseteq Q$ is the set of final states,
    \item $\Sigma_a$ and $\Sigma_b$ are finite sets, corresponding to the input and output alphabets, respectively,
    \item $\delta$ is the state transition function mapping $Q\times(\Sigma_a\cup\{\epsilon\})\to 2^Q$, ie. mapping pairs of states and inputs to the powerset of $Q$,
    \item $\sigma$ is the output function mapping $Q\times (\Sigma_a\cup\{\epsilon\})\times Q\to \Sigma_b^*$, ie. mapping triples of start states, inputs, and end states to arbitrary strings made from the alphabet $\Sigma_b$.
\footnotetext{Some authors define FSTs to have multiple initial states. In practice, this can be equivalently achieved by adding epsilon transitions from $q_0$ to all other desired initial states. Final states can equivalently be represented as having epsilon transitions to a unique final state, which is seen in FSA libraries such as Python's \texttt{k2} library.}
\end{itemize}

The end result maps strings in $\Sigma_a^*$ to a set of strings in $\Sigma_b^*$. Whereas FSAs are equivalent to the set of regular languages, FSTs, by contrast, are equivalent to the set of rational relations between two regular languages. The rational relations between a pair of regular languages form a strict subset of commonly-implemented regular expression substitutions because most implementations include some form of backreferences, which produce patterns that cannot necessarily be expressed in a context-free manner \cite{AHO1990255}. For example, the regex \texttt{/(a*)/} and substitution string \texttt{"{\textbackslash}1b{\textbackslash}1"} cannot be represented with an FST because the contents of capture group \texttt{{\textbackslash}1} may be arbitrarily large, and would thus require an infinite number of states to correctly output the capture group twice. If we use a bounded closure operation rather than Kleene star closure, such as \texttt{/(a\{0,100\})/}, then we can represent the substitution \texttt{"{\textbackslash}1b{\textbackslash}1"} with an FST, albeit with a representation that is less compact than the one afforded by regex with backreferences.

Note that the terms `string' and `alphabet' need not refer to strings of text: FSTs used in ASR systems use an abstract input alphabet consisting of language model outputs (often in phoneme space), and an output alphabet consisting of the decoder's token vocabulary \cite{miao2015eesen}. Such an FST could transduce sequences of phonemes to sequences of words or to sequences of BPE tokens, etc., rather than just characters to characters.

\begin{figure}
    \centering 
    \begin{tikzpicture}[>=stealth',shorten >=1pt,auto,node distance=3 cm, scale = 1, transform shape]

\node[initial,state] (A)                                    {$q_0$};
\node[state]         (B) [right of=A]                       {$q_1$};
\node[state,accepting]         (C) [right of=B]                       {$q_2/0$};

\path[->]
      (A) edge [bend left]      node [align=center]  {$\mathrm{b}:\mathrm{m}/2$} (B)
      (A) edge [bend right]      node [align=center]  {$\mathrm{b}:\mathrm{n}/1$} (B)
      (B) edge [above]      node [align=center]  {$\mathrm{a}:\mathrm{o}/1$} (C)
      (C) edge [loop above] node [align=center]  {$\mathrm{a}:\mathrm{o}/0$} (C);

    \end{tikzpicture}
    \caption{A simple WFST that accepts all strings of the form \texttt{baa*}. For the input string ``baa'', it outputs ``moo'' with weight 3 and ``noo'' with weight 2.}
    \label{fig:wfst}
\end{figure}
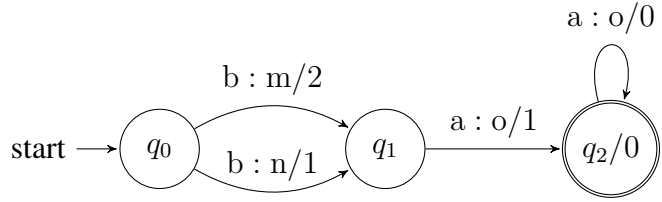

A common generalization of the FST framework is the Weighted Finite-State Transducer (WFST), which assigns weights in a given semiring $K=\langle \mathbb{K}, \oplus,\otimes,\overline{0},\overline{1}\rangle$ to each arc and each final state. We can accommodate this by replacing the output function $\sigma$ with a weighted output relation $\sigma_w\subseteq Q\times(\Sigma_a\cup\{\epsilon\})\times\Sigma_b^*\times Q\times \mathbb{K}$, and adding a final weight function $\Phi:F\to\mathbb{K}$. We can then take the overall weight of a path $\pi$ from $q_0$ to $q_f\in F$ with arcs $[t_1,...,t_n]$ to be \[w(\pi)=\left(\bigotimes_{i=1}^n w(t_i)\right)\otimes \Phi(q_f)\]
where $t_i\in \sigma_w$ and $w(t_i)\in \mathbb{K}$ is the weight associated with $t_i$ for all $i\in [n]$. A pair of input and output strings $\mathbf{x}\in \Sigma_a^*$ and $\mathbf{y}\in \Sigma_b^*$ is accepted with a weight given by $\bigoplus_{\pi} w(\pi)$, over all accepting paths $\pi$ that transduce $\mathbf{x}$ to $\mathbf{y}$ \cite{mohri2002weighted}.

Applications in ASR and other language modeling tasks typically use the weights in an FST as a measure of probability, which naturally lends itself to the \textbf{probability semiring} over the reals, namely $K=\langle\mathbb{R},+,\times,0,1\rangle$. However, to maintain numerical stability with floating-point representations of vanishingly small probabilities, we can store log probabilities spanning $[-\infty,0]$ instead of raw probabilities bound to $[0,1]$, which gives us the \textbf{log semiring} $K=\langle \mathbb{R}_-\cup\{-\infty\}, LSE, +, -\infty, 0 \rangle$, where $LSE$ denotes the Log-Sum-Exp function $LSE(x_1,\dots,x_n)=\log\left(\sum_{i=1}^n e^{x_i}\right)$. An additional caveat is that we often want to identify the best path for some input string $\mathbf{x}$, so rather than adding all possible weights over licensed paths, we want to take the maximum. We can achieve this by using the \textbf{tropical semiring}, which is defined as either the max-plus semiring $K=\langle \mathbb{R}_-\cup\{-\infty\}, \max, +, -\infty, 0 \rangle$ for log probabilities or the min-plus semiring $K=\langle \mathbb{R}_+\cup\{\infty\}, \min, +, \infty, 0 \rangle$ for negative-log probabilities\footnote{These two formulations are isomorphic under negation, and the term ``tropical semiring'' is used alternatively to describe both of them in the literature. In particular, \texttt{rustfst} uses the min-plus formulation and \texttt{k2} uses the max-plus formulation, but both libraries refer to their implementation simply as the ``tropical semiring''.}. This way, while the weight of any one path $\pi$ is still the expected sum of log probabilities, the weight of accepting a string $\mathbf{x}$ becomes $\max_\pi w(\pi)$, which allows us to use pathfinding algorithms such as the Viterbi algorithm to find the best output $\mathbf{y}$ for a given input $\mathbf{x}$.

With the scale of the WFSTs used in language processing, computing an exact solution becomes computationally intractable. The Viterbi algorithm, for instance, takes $O(n\cdot(|Q|+|E|))$ time for an input string of length $n$ and an FST with state set $Q$ and arc set $E$. Since $|E|$ scales quadratically with respect to $|Q|$ in the worst case, this can become intractable even for relatively short inputs for sufficiently dense FSTs with a large number of states. A common optimization to address this is \textbf{beam search}, a pruning method that only considers paths with the best scores for each time step, preserving only the top $k$ states $q$ with the best scores according to a scoring function $\alpha(q)$, where $\alpha$ is computed from $q_0$ up to $q$. An alternate formulation preserves all states $q$ for which $\alpha(q)\ge c\cdot \alpha^*(q)$, where $\alpha^*(q)$ is some optimal score heuristic and $c$ is a constant \cite{ljolje1999efficient}. The former method with a fixed beam width ensures that at most $k$ new states are explored at each timestep, considering the best paths according to local heuristics and potentially sacrificing a globally optimal path in order to find an approximately-optimal path in $O(k^2n)$ time. The latter method, while lacking formal guarantees without properly tuned heuristics, has been found to be effective in practice, especially when used with heuristics that take into account the scores from the current state to a final state and can thus approximate global optimality.

Based on the above, we decided to consider WFSTs in the tropical semiring and perform decoding based on beam search and the Viterbi algorithm. Considering the phonotactics of the language we're working with (discussed below in section \ref{mixtec:phon}), it will be possible to use FSTs at every stage to represent all the replacements with a bounded-length expressions.

Lattice rescoring is a method of integrating neural LM scores into the weights of a WFST. Given a latice--a graph of the hypothesis space for a given input--generated from a WFST, we can dynamically compute neural LM scores for each arc. The key issue is that the number of possible paths, and by extension the number of possible outputs, may be exponential with respect to the number of states, which causes difficulty when trying to run computationally expensive operations over the output space, such as running a neural LM \cite{li2021lattice}. We implement a variation of lattice rescoring in which the log probability scores from a language model are added to our scoring function during beam search decoding. Since the algorithm we implement already computes the best output string on the fly, we do not perform full rescoring of the resultant FST in order to maintain efficiency, although such a construction is feasible if all pruned paths receive weights of $-\infty$.

Another important tool we will utilize is the Mohri-Sproat algorithm for context-sensitive string replacement. Given a replacement rule of the form \mbox{$\mathbf{\phi} \to \mathbf{\psi}\,/\,\mathbf{\lambda}\ \_\ \mathbf{\rho}$}, for regular expressions $\mathbf{\phi},\mathbf{\psi},\mathbf{\lambda},\mathbf{\rho}\in\Sigma^*$, this algorithm constructs five FSTs to insert context markers, perform the replacement, and subsequently remove the markers in an efficient manner \cite{mohri1996efficientcompilerweightedrewrite}. The transducers, in order of composition, are
\begin{itemize}
    \item $r:\Sigma^*\to (\Sigma\cup \{>_\#\})^*$ which inserts the marker $>_\#$ before every instance of $\mathbf{\rho}$ in the input string,
    \item $f:(\Sigma\cup \{>_\#\})^*\to(\Sigma\cup\{>_\#,<_1,<_2\})^*$ which inserts the markers $<_1$ and $<_2$ consecutively before each instance of $\mathbf{\phi}>_\#$ in the input string,
    \item $R_{\mathbf{\phi}\times \mathbf{\psi}}:(\Sigma\cup\{>_\#,<_1,<_2\})^*\to(\Sigma\cup\{<_1,<_2\})^*$ which (1) replaces $\mathbf{\phi}$ with $\mathbf{\psi}$ in the context $<_1 \mathbf{\phi} >_\#$, (2) deletes any instances of $>_\#,<_1,<_2$ that occur within a matching string $\mathbf{\phi}$, and (3) deletes all instances of $>_\#$ that appear in the string (see Figure~\ref{fig:mohri-sproat-fst}),
    \item $\ell_1:(\Sigma\cup\{<_1,<_2\})^*\to(\Sigma\cup\{<_2\})^*$ which only accepts strings in which all instances of $<_1$ occur after $\mathbf{\lambda}$, and deletes $<_1$ from these strings,
    \item  $\ell_2:(\Sigma\cup\{<_2\})\to \Sigma^*$ which only accepts strings in which all instances of $<_2$ do \textit{not} occur after $\mathbf{\lambda}$, and deletes $<_2$ from these strings.
\end{itemize}
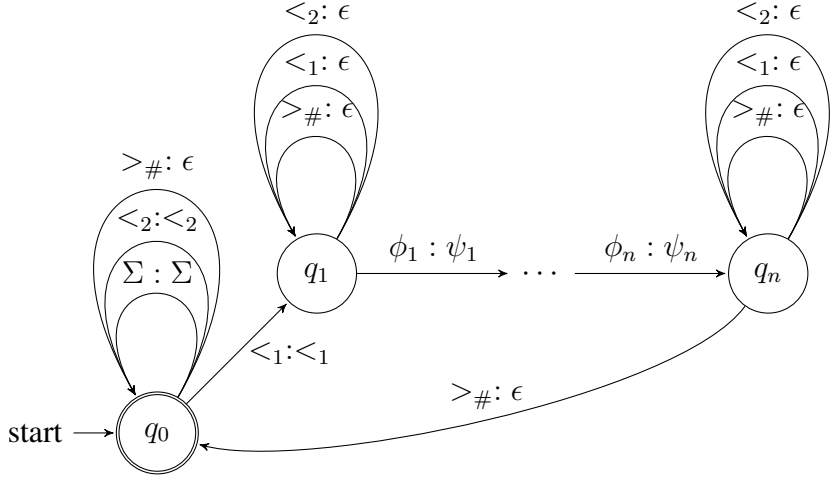
\begin{figure}
    \centering
    \begin{tikzpicture}[>=stealth',shorten >=1pt,auto,node distance=3 cm, scale = 1, ,transform shape]

\node[initial,state,accepting] (A)                                    {$q_0$};
\node[state]         (B) [above right of=A]                       {$q_1$};
\node         (null) [right of=B]                       {$\cdots$};
\node[state]         (C) [right of=null]                       {$q_n$};

\path[->]
      (A) edge [loop above, out=60, in=120, looseness=10]      node [align=center]  {$\Sigma:\Sigma$} (A)
      (A) edge [loop above, out=60, in=120, looseness=15]      node [align=center]  {$<_2:<_2$} (A)
      (A) edge [loop above, out=60, in=120, looseness=20]      node [align=center]  {$>_\#:\epsilon$} (A)
      (A) edge [right]      node [align=center]  {$<_1:<_1$} (B)
      (B) edge [loop above, out=60, in=120, looseness=10]      node [align=center]  {$>_\#:\epsilon$} (B)
      (B) edge [loop above, out=60, in=120, looseness=15]      node [align=center]  {$<_1:\epsilon$} (B)
      (B) edge [loop above, out=60, in=120, looseness=20]      node [align=center]  {$<_2:\epsilon$} (B)
      (B) edge [above]      node [align=center]  {$\phi_1:\psi_1$} (null)
      (null) edge [above]      node [align=center]  {$\phi_n:\psi_n$} (C)
      (C) edge [loop above, out=60, in=120, looseness=10]      node [align=center]  {$>_\#:\epsilon$} (C)
      (C) edge [loop above, out=60, in=120, looseness=15]      node [align=center]  {$<_1:\epsilon$} (C)
      (C) edge [loop above, out=60, in=120, looseness=20]      node [align=center]  {$<_2:\epsilon$} (C)
      (C) edge [bend left, above, out=40, looseness=0.5]      node [align=center]  {$>_\#:\epsilon$} (A)
      ;

    \end{tikzpicture}
    \caption{An illustration of the unweighted machine $R_{\mathbf{\phi}\times\mathbf{\psi}}$ from the Mohri-Sproat construction. Here, $\Sigma:\Sigma$ denotes any character $c\in\Sigma$ being transduced to itself, and $\phi_1\dots\phi_n$ and $\psi_1\dots\psi_n$ are the respective expansions of $\mathbf{\phi}$ and $\mathbf{\psi}$, normalized to have the same length for simplicity.}
    \label{fig:mohri-sproat-fst}
\end{figure}

Each of these automata can be constructed in linear time and space with respect to the number of states in a deterministic FST representing $\mathbf{\lambda},\mathbf{\rho},\mathbf{\phi}$, and $\mathbf{\psi}$ \cite{mohri1996efficientcompilerweightedrewrite}. The overall replacement FST is formed by the composition of these machines, \mbox{$\mathcal{T}_{repl}=r\circ f \circ R_{\mathbf{\phi}\times\mathbf{\psi}} \circ \ell_1 \circ \ell_2$}. FST composition is defined similarly to function composition: for some pair of machines $f:\Sigma_a^*\to\Sigma_b^*, g: \Sigma_b^*\to\Sigma_c^*$, then for all $\mathbf{x}\in \Sigma_a^*,\mathbf{y}\in\Sigma_b^*,\mathbf{z}\in\Sigma_c^*$ where $f$ transduces $\mathbf{x}\to\mathbf{y}$ and $g$ transduces $\mathbf{y}\to\mathbf{z}$, the composition $f\circ g:\Sigma_a^*\to\Sigma_c^*$ transduces $\mathbf{x}\to\mathbf{z}$. The FSTs in the Mohri-Sproat replacement algorithm can also be weighted, allowing us to model replacement rules that are probabilistic rather than obligatory. In this case, note that the weight of a composed path is equal to the product of paths from each WFST in the appropriate semiring. That is, if machine $f$ has paths $\pi_f(\mathbf{x},\mathbf{y})$ and $g$ has paths $\pi_g(\mathbf{y},\mathbf{z})$ for some $\mathbf{x}\in \Sigma_a^*,\mathbf{y}\in\Sigma_b^*,\mathbf{z}\in\Sigma_c^*$, then the weight of transducing $\mathbf{x}\to\mathbf{z}$ in $f\circ g$ is
\[\bigoplus_\mathbf{y} w(\pi_f(\mathbf{x},\mathbf{y}))\otimes w(\pi_g(\mathbf{y},\mathbf{z}))\]

Replacement rules of the form \mbox{$\mathbf{\phi} \to \mathbf{\psi}\,/\,\mathbf{\lambda}\ \_\ \mathbf{\rho}$} are useful in various areas of natural language processing, such as handling orthographic rules and sound changes \cite{mortensen2018epitran}. Since our project involves context-sensitive text replacement as part of our FST pipeline, we implement a version of this algorithm for our experiments.

\chapter{Yolox\'ochitl Mixtec}

Yolox\'ochitl Mixtec (YM) is an endangered language spoken in southern Mexico. YM belongs to the Mixtec language family, which includes a wide variety of languages spoken in the states of Oaxaca, Guerrero, and Puebla. YM is spoken in four villages in the state of Guerrero--Yolox\'ochitl, Arroyo Cumiapa, Cuanacaxtitlan, and Buenavista--with roughly 10,000 speakers as of 2016 \cite{palancar201612}. Almost all speakers in the first two villages are highly fluent in YM, while in the latter two, YM is rapidly disappearing as the younger generation increasingly switches to Spanish.

{Like other Mixtecan languages,} YM has phonemic tones and tonal morphology. There are nine basic tones in total: four level tones, written as numbers $\langle 1\rangle$ (IPA: \textipa{/\tone{11}/}) to $\langle 4\rangle$ (IPA: \textipa{/\tone{55}/}); three rising tones, $\langle 13\rangle$, $\langle14\rangle$, and $\langle24\rangle$ (IPA: \textipa{/\tone{14}/, /\tone{15}/, /\tone{25}/}); and two falling tones, $\langle 32\rangle$ and $\langle 42\rangle$ (IPA: \textipa{/\tone{42}/, /\tone{52}/}). Additionally, a handful of additional tones appear in the corpus, including some infrequent contour tones (namely $\langle 143 \rangle$, $\langle132\rangle$, and $\langle342\rangle$), the falling tones $\langle41\rangle$ and $\langle43\rangle$ that mainly appear in filler words, and a rising tone, $\langle34\rangle$, which seems to occur only as a result of a morphological process.

Tonal morphology--a type of non-concatenative morphology that affects phonemic tones--is used in YM to mark both derivational and inflectional properties, with regular patterns seen across the language. These tonal processes must be accounted for when building a tokenization schema that is morphologically robust, as these morphemes play a central role in the language.
Additionally, in any language, an analysis of the morphology should be prefaced by an overview of the phonotactic structure, as the latter informs the licensed structures and forms that the morphology may take.
\section{Phonotactics}
\label{mixtec:phon}

\begin{table}[b!]
    
    \centering
    \hspace*{-2cm}
    \begin{tabular}{r|ccccc}
         \textbf{Tone}\; & \multicolumn{5}{c}{\textbf{Tone on }$\mathbf{\mu_1}$} \\
         \textbf{on }$\mathbf{\mu_2}$ & \mapcolumn{\orth}{1}{3}{4}{13}{14} \\
         \hline
         \orth{1} & \langstack{bi1ka1}{comb} & -- & \langstack{ya'4a1}{brown} & -- & \langstack{na'14a1}{demoniac}
         \\
         \orth{2} & -- & \langstack{\~nu3u2}{village} & \langstack{nda4a2}{where} & \langstack{kwe13e2}{linger} & \langstack{ndi14i2}{pink}
         \\
         \orth{3} & \langstack{ta1a3}{man} & \langstack{xa3a3}{fast} & \langstack{i'4in3}{mute} & \langstack{nda13a3}{went up} & \langstack{nu14u3}{face}
         \\
         \orth{4} & \langstack{xi1i4}{grandfather} & \langstack{bi3ko4}{feast} & \langstack{nda4a4}{black} & \langstack{tu13u4}{stripped} & \langstack{ye'14e4}{door}
         \\
         \orth{13} & -- & \langstack{kia'3bi13${\color{black}{}^*}$}{to be sold} & \langstack{che4e13}{big} & \langstack{kia'13bi13${\color{black}{}^*}$}{was sold} & \langstack{ndo14o13}{to not stay}
         \\
         \orth{14} & -- & \langstack{i3nda14${\color{black}{}^*}$}{one} & \langstack{nu4na14}{habitually open} & -- & \langstack{ke14yu14}{to not swim}
         \\
         \orth{24} & -- & -- & \langstack{ya4a24}{tongue} & -- & \langstack{ka14a24}{to not slip}
         \\
         \orth{32} & \langstack{xa1ko32}{opossum} & -- & -- & -- & \langstack{ku14xi32${\color{black}{}^*}$}{to not rust}
         \\
         \orth{42} & \langstack{ta1kwi42}{water} & \langstack{\~nu3u42}{night} & \langstack{chi4lin42}{lark song} & -- & --
         \\
    \end{tabular}
    \hspace*{-2cm}
    \caption{A table of the attested tonal patterns in couplets in YM \cite{palancar201612,dicanio2014phonetics}. Entries marked with an * are only attested in one word. Derived and inflected forms are included.}
    \label{tab:tone_dist_1}
\end{table}

The phonotactic structure of YM lends itself to analysis in terms of morae rather than syllables. A mora is a unit of length, which in YM corresponds to a tone-bearing unit. A single syllable in YM can span more than one mora, for instance $\langle$\olang{\~nu1u42}$\rangle$ is a single syllable but has two tone markers, creating a tone pattern \orth{1-42}, which is not attested as a monomoraic tone.
Most Mixtec roots are bimoraic, carrying two tones in the overall melody in a structure commonly referred to as a ``couplet'' \cite{dicanio2014phonetics,dicanio2020phonetic}. These couplets include both the monosyllabic CV(\?)V structure as well as the disyllabic CV(\?)CV structure. Many trimoraic roots are also attested, having a structure of CVCV(\?)V or  CVCV(\?)CV \cite{dicanio2020phonetic}. A handful of quadrimoraic roots seem to appear in the corpus, but these account for less than 70 unique types and have not been studied in detail.

The number of attested bimoraic tonal melodies is smaller than what would be predicted if any of the basic tones could appear on each mora. Out of the nine basic tones, only five--\orth{1}, \orth{3}, \orth{4}, \orth{13}, and \orth{14}--are licensed to appear on $\mu_1$, the first mora of a couplet, while all are licensed to appear on the second mora, $\mu_2$. Table~\ref{tab:tone_dist_1} summarizes the 31 attested tone melodies that appear across the lexicon. Of these, only 14 appear in the irrealis form of a bimoraic verb stem, as summarized in Table~\ref{tab:tone_dist_2}.

\begin{table}[]
    \centering
    
    \begin{tabular}{r|ccccc}
     & \multicolumn{5}{c}{$\mu_1$} \\
    $\mu_2$ & \mapcolumn{\orth}{1}{3}{4}{13}{14} \\
    \hline
    \orth{1} & 45 & -- & 1 & -- & 1 \\
    \orth{2} & -- & 25 & 0 & 0 & 7 \\
    \orth{3} & 29 & 79 & 0 & 0 & 10 \\
    \orth{4} & 29 & 46 & 0 & 0 & 1 \\
    \orth{13} & -- & 1 & 1 & 0 & 0 \\
    \orth{14} & -- & 0 & 0 & -- & 0 \\
    \orth{24} & -- & -- & 0 & -- & 0 \\
    \orth{32} & 2 & -- & -- & -- & 0 \\
    \orth{42} & 0 & 0 & 0 & -- & -- \\
    \end{tabular}
    \caption{Occurrence rates of tonal melodies in bimoraic lexical verb stems found in Amith's dictionary \cite{amithdictionary}. As convention, the lemma is taken to be the irrealis form. An entry of `0' indicates an attested tonal melody that does not appear in irrealis verbs but is attested elsewhere, whereas a `--' indicates a melody that is not attested anywhere in the lexicon.}
    \label{tab:tone_dist_2}
\end{table}

\section{Morphology}
As demonstrated in Table~\ref{tab:tone_dist_2}, many tone melodies are either infrequent (eg. \orth{14-4}) or completely unattested (eg. \orth{13-2}) in the irrealis form of bimoraic verbs, which we take to be the lemma. However, these melodies are attested in inflected forms of verbs, with 66 non-lemma appearances of \orth{14-4} and 32 appearances of \orth{13-2}. Several of these are a result of the tonal morphology marking for aspect and mood on verbs.

Verbs in YM inflect for completive and habitual aspect (glossed as \acrshort{cpl} and \acrshort{hab}, respectively), the irrealis mood (glossed as \acrshort{irr}), and negation (glossed as \acrshort{neg}). We take the positive irrealis form of each verb as the lemma.

The completive has two alternative forms: the prefix \orth{\olang{ni1-}}, and the insertion of a low tone to $\mu_1$ of the root. This holds for both bimoraic and trimoraic roots, as shown in (\ref{sent:cpl-bimora}) and (\ref{sent:cpl-trimora}). These forms are noted to be in free variation with each other \cite{palancar201612}. Notably, the tonal form of the completive aspect on lemmas with tone \orth{1} or \orth{14} on $\mu_1$ is homophonous with the respective irrealis form.

\eenumsentence{
\label{sent:cpl-all}
\begin{multicols}{3}
\item
\label{sent:cpl-bimora}
    \shortex{1}{\ggglang{chi3chin4}}{suckle.\acrshort{irr}}{\tlang{`to suckle'}}
    \shortex{2}{\ggglang{ni1-}&\ggglang{chi3chin4}}{\acrshort{cpl}-&suckle}{\tlang{`suckled'}}
    \shortex{1}{\ggglang{chi\{>1\}3chin4}}{suckle\{\acrshort{cpl}\}}{\tlang{`suckled'}}
\end{multicols}

\begin{multicols}{3}
\item
\label{sent:cpl-trimora}
    \shortex{1}{\ggglang{chi3nda'3a4}}{push.\acrshort{irr}}{\tlang{`to push'}}
    \shortex{2}{\ggglang{ni1-}&\ggglang{chi3nda'3a4}}{\acrshort{cpl}-&push}{\tlang{`pushed'}}
    \shortex{1}{\ggglang{chi\{>1\}3nda'3a4}}{push\{\acrshort{cpl}\}}{\tlang{`pushed'}}
\end{multicols}
}

The habitual aspect is marked by overwriting the lexical tone on $\mu_1$ with tone \orth{4}. However, some additional patterns can be observed. Notably, some melodies push the lexical tone from $\mu_1$ onto $\mu_2$, such as \orth{1-3} in the irrealis becoming \orth{4-13} for the habitual. The tonal allomorphs of the habitual aspect are summarized in Table~\ref{tab:habitual-allomorphy}. Bimoraic verbs with tone \orth{14} in $\mu_1$ mark the habitual aspect with the prefix \orth{\olang{i4-}} rather than a tonal overwrite \cite{palancar201612}, as exemplified by (\ref{sent:hab14}).
\begin{table}[t!]
    \centering
    
    \begin{tabular}{c|c|cc}
        \textbf{Structure} & \textbf{Melody} & \textbf{Segmentation} & \textbf{Gloss} \\
        \hline
        CVCV & 1-1 & \ggglang{ki\gggsub{1}{4}xin1} & fall\_asleep\{\acrshort{hab}\} \\
        CVV  & 1-1 & \ggglang{tu\gggsub{1}{4}un1} & catch\_fire\{\acrshort{hab}\} \\
        \hline
        CVCV & 1-3 & \ggglang{ka\gggsub{1}{4}ku\gggsub{}{1}3} & escape\{\acrshort{hab};1,2\} \\
        CVV  & 1-3 & \ggglang{ka\gggsub{1}{4}an\gggsub{}{1}3} & get\_accustomed\{\acrshort{hab};1,2\} \\
        \hline
        CVCV & 1-4 & \ggglang{ka\gggsub{1}{4}xan\gggsub{}{1}4} & sneeze\{\acrshort{hab};1,2\} \\
        CVV  & 1-4 & \ggglang{ku\gggsub{1}{4}un\gggsub{}{1}4} & be\_ground\{\acrshort{hab};1,2\} \\
        \hline
        CVCV & 3-3 & \ggglang{ka\gggsub{3}{4}ba3} & sleep\{\acrshort{hab}\} \\
        CVV  & 3-3 & \ggglang{chi\gggsub{3}{4}i\gggsub{3}{4}} & get\_wet\{\acrshort{hab};1,2\} \\
        \hline
        CVCV & 3-4 & \ggglang{ku\gggsub{3}{4}chi4} & feel\_sad\{\acrshort{hab}\} \\
        CVV  & 3-4 & \ggglang{ka\gggsub{3}{4}a\gggsub{}{2}4} & slip\{\acrshort{hab};1,2\} \\
        \hline
        CVCV & 14-X & \ggglang{i4- chu14tu2} & \acrshort{hab}- kiss \\
        CVV  & 14-X & \ggglang{i4- xio'14o4} & \acrshort{hab}- get\_sick\_from\_craving \\
    \end{tabular}
    \caption{Summary of inflectional classes for the habitual aspect on bimoraic verbs.}
    \label{tab:habitual-allomorphy}
\end{table}

\enumsentence{
\label{sent:hab14}
\shortex{2}{\mapcolumn{\ggglang}{i4-}{chu14tu2}}{\acrshort{hab}- & kiss}{\tlang{`habitually kiss'}}
}

Negation is most commonly realized by overwriting the lexical tone in $\mu_1$ with \orth{14}. This occurs for the irrealis forms of all verbs, and is licensed to mark negation for the completive along with the prefix \orth{\olang{ni1-}}. For irrealis stems with tone \orth{14} in $\mu_1$, the negative adverb \orth{\olang{kwa14}} is used to explicitly mark negation. Some such verbs also have an alternative form with \orth{4} in $\mu_1$ to mark the negative, with or without the adverb \orth{\olang{kwa14}}. The adverb \orth{\olang{ba143}} is also used to form negatives, particularly for the habitual form and the completive form. Note that when \orth{\olang{ba143}} is used with \orth{\olang{ni1-}}, the completive prefix is raised to a high tone \orth{4}\footnote{The prefix \orth{\olang{ni4-}} may also denote the counterfactual mood, but this is not confirmed at the time of writing}. A summary of licensed negative forms for two verbs is presented in (\ref{sent:neg-all}).

\eenumsentence{
\label{sent:neg-all}
\begin{multicols}{2}
\item
\shortex{1}{\ggglang{cho'\gggsub{3}{14}ma4}}{squash\{\acrshort{neg}.\acrshort{irr}\}}{\tlang{`to not squash'}}

\shortex{2}{\mapcolumn{\ggglang}{ni1\gggsub{}{4}-}{cho'3ma4}}{\acrshort{cpl}\{\acrshort{neg}\} & squash}{\tlang{`didn't squash'}}

\shortex{3}{\mapcolumn{\ggglang}{ba143}{ni\gggsub{1}{4}-}{cho'3ma4}}{\acrshort{neg} & \acrshort{cpl}\{\acrshort{neg}\} & squash}{\tlang{`didn't squash'}}

\shortex{2}{\mapcolumn{\ggglang}{ba143}{cho'\gggsub{3}{4}ma4}}{\acrshort{neg} & squash\{\acrshort{hab}\}}{\tlang{`usually doesn't squash'}}
\end{multicols}

\begin{multicols}{2}
\item
\shortex{2}{\mapcolumn{\ggglang}{kwa14}{xi14ko3}}{\acrshort{neg}.\acrshort{irr} & sell}{\tlang{`to not sell'}}

\shortex{2}{\mapcolumn{\ggglang}{(kwa14)}{xi\gggsub{1}{}4ko3}}{(\acrshort{neg}.\acrshort{irr}) & sell\{\acrshort{neg}.\acrshort{irr}\}}{\tlang{`to not sell'}}

\shortex{2}{\mapcolumn{\ggglang}{ni1\gggsub{}{4}-}{xi14ko3}}{\acrshort{cpl}\{\acrshort{neg}\} & sell}{\tlang{`didn't sell'}}

\shortex{3}{\mapcolumn{\ggglang}{ba143}{i4-}{xi14ko3}}{\acrshort{neg} & \acrshort{hab}- & sell}{\tlang{`usually doesn't sell'}}
\end{multicols}
}

Trimoraic roots exhibit the same $\mu_1$ tone overwrites as bimoraic verbs, albeit with almost no pushing of lexical tones onto subsequent moras. The only case where this is observed is in the verb \orth{\olang{ke1nu3u3}}, which has habitual and negative forms as shown in example (\ref{sent:kenuu}). The three quadrimoraic roots listed in the dictionary also exhibit no pushing of lexical tones after the initial overwrite, so we will treat this occurrence as an irregularity among verbs longer than two moras \cite{amithdictionary}.

\eenumsentence{
\label{sent:kenuu}
\begin{multicols}{2}
    \item 
    \shortex{1}{\ggglang{ke\gggsub{1}{4}nu\gggsub{3}{1}u3}}{slide\_down\{\acrshort{hab};1,2\}}{\tlang{`often slides down'}}
    
    \item
    \shortex{1}{\ggglang{ke1\gggsub{}{4}nu\gggsub{3}{1}u3}}{slide\_down\{\acrshort{neg};1,2\}}{\tlang{`doesn't slide down'}}
\end{multicols}
}

In addition to the inflectional morphemes discussed above, a handful of derivational morphemes are also marked with tone shifts. One of these is a detransitive marker, which makes an intransitive verb out of a transitive verb, and is marked by replacing tone \orth{3} with tone \orth{1} in $\mu_1$, as shown in example (\ref{sent:detrans}). Note that since the derivation occurs before inflection, further inflection will pattern with a root that has tone \orth{1} in the first mora. The chain of processes in the first mora that may be generated from this is marked explicitly with a multi-arrow notation, such as \ggglang{\gggsub{3}{1>4}}.
The other major tonal derivational morpheme is a shift from tone \orth{1} to tone \orth{4} in $\mu_1$, which forms an adjective from a noun, as in example (\ref{sent:adj}).

\eenumsentence{
\label{sent:detrans}
\begin{multicols}{2}
    \item
    \shortex{1}{\ggglang{cho'\gggsub{3}{1}ma4}}{crush\{\acrshort{detr}\}}{\tlang{`to crush'}}
    \item
    \shortex{1}{\ggglang{cho'\gggsub{3}{1>4}ma\gggsub{}{1}4}}{crush\{\acrshort{detr};1\}\{\acrshort{hab};1,2\}}{\tlang{`often crushes'}}
\end{multicols}
}

\enumsentence{
\label{sent:adj}
\shortex{1}{\ggglang{yu\gggsub{1}{4}u4}}{stone\{\acrshort{n>adj}\}}{\tlang{`solid'}}
}

Subject markers for verbs and possession markers for nouns are marked with suffixing enclitics in YM. Some, such as the second person plural enclitic \orth{\olang{=ndo4}}, only have one allomorph, but most have several allomorphs that are phonologically conditioned. For instance, the first person singular enclitic takes the form \orth{\olang{=e1}} following a stem with tone \orth{1} in the final mora and ending with \orth{a} or \orth{o}, \orth{\olang{=i1}} for stems ending with \orth{u} and tone \orth{1}, \orth{\olang{=yu1}} for all other stems ending with tones \orth{1} or \orth{2}, and \orth{\olang{=2}} for stems ending with tones \orth{3} or \orth{4}. Notably, the allomorph \orth{\olang{=2}} is a strictly tonal morpheme, which may trigger elision of the underlying stem-final tone depending on the melody of the stem. However, it is transcribed as an enclitic in the practical orthography of YM for consistency with the other allomorphs. %

The tonal morphology of YM exhibits complexity that is not easily captured by traditional tokenization methods. We intend to build tokenization schemes that split tones and segments in hopes of producing a token vocabulary that is smaller and more meaningful than can be produced with BPE alone.

\chapter{Methods}

\section{ASR Task}

Previous work on YM has achieved very good performance in ASR using the practical orthography, with the work by Shi et al. achieving a CER of 7.7\% for an end-to-end conformer-based model with a Unigram-based\footnotemark tokenizer \cite{shi2021leveragingendtoendasrendangered}. Shi et al. also experimented with end-to-end transformer models, obtaining a CER of 7.9\%, only slightly behind the conformer model. Tokenizing by morae on the conformer model resulted in a slightly higher CER of 9.4\%, while still outperforming naive word-based and morpheme-based tokenization \cite{amith2021end}.
The model with the best performance was built from a split of a corpus with 92 hours of annotated audio, although a model trained on only 50 hours still achieved a CER of 8.7\% on the test split, which is within the 10\% guideline outlined by Amith et al. in \cite{amith2021end}.
\footnotetext{ESPnet supports both BPE and Unigram tokenizers with the SentencePiece library, but labels both of them as ``modes'' of BPE. Shi refers to the tokenizer as BPE in his papers, but notes that it is constructed from a Unigram model. We will refer to this as a Unigram tokenizer for clarity.}

We will utilize the same basic E2E conformer and transformer architectures as developed by Shi et al., with updates focused on supporting various novel tokenization methods and adapting to the updated dataset.
It is worth noting that while Shi's original experiments were run on \mbox{ESPnet1}, we will be using \mbox{ESPnet2} to run the recipes for our experiments, and as such, we will be considering a few different configurations. In addition to the default configurations for conformer and transformer, we will also consider some variations from the default transformer. In particular, we will test adding SpecAugment\cite{Park-2019-specaugment} and replacing Chainer initialization with Xavier Uniform initialization in the transformer configuration--both of these changes are already implemented in the conformer configuration out of the box. Additionally, per Shi's recommendation, I also tested different subsampling factors for the transformer model's input layer.

Performance on the ASR task will be measured with character error rate (CER) and word error rate (WER), and will be compared against the unsegmented reference text. We will compare all models against a Unigram baseline, and run the remainder of experiments against the model with the best performance for Unigram.

\section{Tokenization Schemes}

As a baseline, we will run the ASR experiments on the new dataset with a BPE tokenizer, a Unigram tokenizer, a word-level tokenizer, and a character-level tokenizer. All of these are natively supported in \mbox{ESPnet}. We will also construct a WordPiece tokenizer using existing infrastructure from the Hugging Face tokenizer library\cite{tokenizers}, which can be used indirectly with ESPnet. Afterwards, we will test two novel tokenization methods: a Segment-and-Melody tokenizer, which extracts the tonal sequences without predicting segmentation, and a Sequence of Processes tokenizer, which predicts tonal processes in a way that segmentation can be produced.
We designed these new tokenizers to address the inadequacies discussed in Section~\ref{bg:tokenization} for processing the non-concatenative morphology of YM.

\subsection{Segment-and-Melody}

The first scheme we will examine splits the text regularly into separate tokens to indicate the segments (ie. non-tonal information in each mora) and the tonal melody. Based on the observed patterns of morae as regular in YM, we can use a simple regular expression to extract segments and their corresponding tones, which we use in constructing tokens.

The tokenization process is as follows:
\begin{itemize}
    \item Strip punctuation and separate enclitics and productive affixes.
    \item For each word, maintain a list of segments and a list of tones.
    \item For each match of the regular expression \texttt{([a-z\~n']+)([1-4]+)} in a word, we append match group 1 to the list of segments and match group 2 to the list of tones.
    \item Append any remaining characters to the segments list, plus an empty string to the tones list.
    \item If the tones list is empty, treat the word as Spanish and add the full word as a token.\footnote{In the standalone Segment-Melody tokenizer, no further subword processing is performed on the Spanish tokens. The augmented SentencePiece variants do apply subword tokenization.}
    \item Otherwise, join the strings in the segments list and the tones list with a pipe ``\texttt{|}''.  Output one token containing the segments, followed by a token containing the tones.
\end{itemize}

As an example, \orth{\olang{ta'14bi4}} from (\ref{sent:tabi-tokenizer}) above would be tokenized as \texttt{ta'|bi~14|4}. This representation would allow a language model to separately learn the distributions of the segments \orth{ta'-bi} and the overall tone melody \orth{14-4}. We found 155 unique attested melody sequences in Amith's dictionary and some 1925 unique segments, yielding a total of roughly 2100 possible tokens, excluding Spanish words and punctuation\cite{amithdictionary}.

Reconstructing a well-formed token sequence is straightforward: simply identify pairs of segmental tokens and tonal tokens and match them up. However, we must account for erroneous outputs from the decoder model, such as when a segmental token and its corresponding tonal token do not have the same length. In this case, we pad the {end} of the tonal token with a placeholder tone, represented by \texttt{\#}, and take the {prefix} of appropriate length to match the number of moras in the segment token.

While it would be ideal to run Asgari et al.'s MorphBPE as an experiment, their formulation requires gold-standard morphological segmentations and definitions of morphological boundaries \cite{asgari2025morphbpemorphoawaretokenizerbridging}. Since we do not yet have this data for YM, we decided to approximate their methods by applying BPE to the output of the Segment and Melody tokenizer. Since this tokenizer separates concatenative affixes and enclitics from their bases and also separates the tone melodies from the rest of the word, we effectively guarantee the desired property of minimizing BPE merges that cross over morpheme boundaries. The only cases where it would be possible for BPE to merge across a morpheme boundary are in cases of segmental overwriting and tone melodies that contain both lexical tones and tonal morphology.

\subsection{Sequence-of-Processes}

In order to represent tonal morphology in a linear manner, we present a tokenization scheme that separates inflected words into a sequence containing the corresponding lemma and the tonal processes in the derivation. The G3 representation allows us to easily segment concatenative morphemes--such as affixes and enclitics--into separate tokens from the root, but handling the non-concatenative processes requires nontrivial processing. Taken as-is, applying subword tokenization with a coarser granularity than character-level to the G3 forms runs the same risk of splitting lexical and morphological information across multiple tokens as does a similar tokenization of the unsegmented forms. However, the moraic structure of YM allows for a way to unambiguously represent the tonal morphology as a sequence following the lemma. Specifically, we mirror the G3 notation, but list the tonal processes for each mora after the word rather than inline. A few modifications need to be made for this to work: first we introduce tokens representing no change, such as \texttt{1>1}, which are not marked in G3. Second, we expand the rewrite rule to cover the whole tonal pattern for the mora, so while \ggglang{chi'\{>1\}3i3} is written as the insertion of a tone 1 before an unchanged tone 3 in G3, our tokenizer would use \texttt{3>13}. Example~(\ref{sent:tabi-tokenizer}) above, \orth{\olang{ta'14bi4}}, would be represented in this scheme as \texttt{ta'3bi4 3>14 4>4}.

We had considered approaches involving tokens that more closely reflected the morphology, but this presented issues for reconstruction: the appearance of a tone \orth{1} at the onset of a mora is licensed in both $\mu_1$ and $\mu_2$, so a token simply encoding the process \orth{\ggglang{\gggsub{}{1}}} without further context is not sufficient to reconstruct the segmentation. Similarly, for tonal melodies such as \orth{3-3} where multiple moras carry the same tone, a token simply encoding the process \orth{\ggglang{\gggsub{3}{4}}} could be applied to either mora. We decided that having one token per mora was an adequate representation of the tonal morphology while allowing for unambiguous reconstruction. Additionally, we decided not to account for the segmental overwriting in this schema because we observed segmental variation in only 99-114\footnote{Palancar, Amith, and Garc\'ia's 2016 paper cites 109 ``variant verbs'' that exhibit these patterns \cite{palancar201612}. Regular expression searches on a version of Amith's dictionary from June 2025 yielded only 99 matches when filtering out any segmental match in the verbs, but as many as 114 when our search criteria were broadened \cite{amithdictionary}. The dictionary data is actively being cleaned, so typos are a likely source of erroneous hits and misses.} unique lexemes, most of which are frequently used verbs. The variation itself is also somewhat irregular: while patterns like \orth{\ggglang{\gggsub{xi}{ku}}} are attested, many variations of the pattern exist, and they are not seen outside the limited set of variant verbs. Based on the glosses of the observed variant verbs, we can expect that these verbs occur frequently within the corpus. As such, we predict the model will have enough examples to pick up on these irregular forms as-is, similarly to how language models are able to interpret irregular apophony in common English words like ``\olang{sat}'' (G3: \ggglang{s\gggsub{i}{a}t}) or ``\olang{geese}'' (G3: \ggglang{g\gggsub{oo}{ee}se}) without explicit representations of the morphology.

A key issue is that our sequence-of-processes tokenization cannot be produced unambiguously from the unsegmented practical orthography. Although many tokens can be unambiguously segmented in practice, there are words like \orth{\olang{i4in4}} in the corpus that have multiple meaningful segmentations, which are shown in (\ref{sent:i4in4-segmentations}a-d). Using the tokenization method described above would only generate three of the four forms, since both (\ref{sent:i4in4-hang}) and (\ref{sent:i4in4-seem}) would be tokenized as \texttt{i3in3 3>4 3>4}, but a context-free model cannot adequately distinguish which of the licensed forms is correct for any specific instance of the word \orth{\olang{i4in4}}. To remedy this, we propose a cascaded system wherein an FST generates a set of licensed forms for each word it sees, and a language model trained on segmented data selects the most likely form in context. The usage of an FST restricts the output space to only include valid token sequences, while the language model allows for selection of an appropriate segmentation in context.

\eenumsentence{
\label{sent:i4in4-segmentations}
\begin{multicols}{2}
\item
\shortex{1}{\ggglang{i\gggsub{1}{4}in4}}{salt\{\textsc{N>Adj}\}}{\tlang{`salty'}}
\item
\label{sent:i4in4-hang}
\shortex{1}{\ggglang{\gggsub{kw}{}i\gggsub{3}{4}in\gggsub{3}{4}}}{hang.\acrshort{pl}\{\acrshort{est};1,2,3\}}{\tlang{`(they) hang it'}}
\item
\label{sent:i4in4-seem}
\shortex{1}{\ggglang{i\gggsub{3}{4}in\gggsub{3}{4}}}{appear.as\{\acrshort{hab};1,2\}}{\tlang{`(they) seem'}}
\item
\shortex{1}{\ggglang{i4in4}}{hail}{\tlang{`A hailstone'}}
\end{multicols}
}

Examples (\ref{sent:i4in4-hail-context}) and (\ref{sent:i4in4-seem-context}) below highlight full sentences where the word \orth{\olang{i4in4}} has different segmentations that our model needs to distinguish. Both segmentations are licensed under the FST, but a language model would be able to recognize that the interpretation as the noun ``\tlang{hail}'' fits better in (\ref{sent:i4in4-hail-context}) where rain is discussed, while the interpretation as the verb ``\tlang{seem}'' fits better in (\ref{sent:i4in4-seem-context}) where the likeness of an object is discussed.

\enumsentence{
\label{sent:i4in4-hail-context}
\shortex{6}{\mapcolumn{\ggglang}{ku\gggsub{3}{4}un\gggsub{3}{4}}{sa1bi4}{ndi4}{nda1}{\textbf{i4in4}}{ko\gggsub{1}{4}yo\gggsub{}{1}3}}{fall\{\acrshort{hab};1,2\} & rain & \acrshort{comp} & even & hail & fall.\acrshort{pl}\{\acrshort{hab};1,2\}}{\tlang{`It would rain so much that it even hailed.'}}
}

\enumsentence{
\label{sent:i4in4-seem-context}
\shortex{9}{\mapcolumn{\ggglang}{na4tu2}{kwi3i4}{=o4}{=run4}{ndi4} & [...] & \mapcolumn{\ggglang}{tan42}{\textbf{i\gggsub{3}{4}in\gggsub{3}{4}}}{i3tun4}...}{if & peel & =1\acrshort{pl}.\acrshort{incl} & =3.wood & \acrshort{comp} &  & when & appear\_as\{\acrshort{hab};1,2\} & tree}{\tlang{`If you peel the wood, it looks like that of this tree...'}}
}

To best leverage both the language model and the FST in this process, we will use lattice rescoring to reweight the FST paths based on the probabilities from the language model for generating each possible sequence. We attempted to use the lattice rescoring functions available in the \texttt{k2} library, but ultimately were unable to directly connect it to our segmentation LM. Instead, we directly implemented some modified versions of the Viterbi algorithm and Beam Search, where the probability heuristic at each step accounts for both the FST weights and the language model probabilities.

The language model we use is a fine-tuned version of ByT5 trained on a transformed version of the text corpus in which all unambiguous words are replaced with their segmentation, and all ambiguous words are prepended by the \texttt{\$} character. In this way, the model can learn character-level morphological patterns for a large number of cases and, ideally, make accurate judgments about the quality of different proposed segmentations. Since the FST doesn't license the \texttt{\$} character as an output, the model will effectively be forced to choose the next-best segmentation in context.

Formally, consider a transducer $\mathcal{T}:\Sigma_a^*\to \Sigma_b^*$ that accepts the string \mbox{$\mathbf{x}=x_0\cdots x_n\in \Sigma_a^*$} and a language model $\mathcal{M}: \Sigma_a^*\to\Sigma_b^*$ that outputs probabilities for a segmentation $\mathbf{y}=y_0\cdots y_m\in \Sigma_b^*$ of the form $p(y_{i+1}\cdots y_{i+k}|\mathbf{x},y_0\cdots y_i)$. For each arc $t_{j+1}=\langle q, x_{j+1}, y_{i+1}\cdots y_{i+k}, q', w(t_{j+1})\rangle$ from $q\to q'$ that transduces $x_{j+1}:y_{i+1}\cdots y_{i+k}$ with weight $w(t_{j+1})$ in $\mathcal{T}$, and prefix strings $x_0\cdots x_j\in \Sigma_a^*, y_0\cdots y_i\in \Sigma_b^*$ that are licensed at state $q$ by following an optimal path $\pi^*(q_0,q)=[t_1,...,t_j]$, we take the weight of reaching that state to be
\begin{align*}
    \hat{w}(q'|q)&=\log p(y_0\cdots y_{i+k}|\mathbf{x})+w(t_{j+1})+\sum_{t\in \pi^*(q_0,q)} w(t)\\
    &=\log p(y_{i+1}\cdots y_{i+k}|\mathbf{x},y_0\cdots y_{i}) + w(t_{j+1}) + \hat{w}(q)
\end{align*}

where the initial state is assigned $\hat{w}(q_0)=0$. Ideally, for each new state $q'$, we want to find the state and arc that maximizes this weight:
\[\hat{w}(q')=\max_{q:\delta(q,x_j)\ni q'} \hat{w}(q'|q)\]
Finally, take $\hat{q}=\arg\max_{q_f\in F} \hat{w}(q_f)$, and take $\mathbf{\hat{y}}=y_0\cdots y_m\in\Sigma_b^*$ to be the output string obtained by following the optimal path $\pi^*(q_0,\hat{q})$. This value, $\mathbf{\hat{y}}$, is our predicted segmentation of the input string $\mathbf{x}$.

In order to maintain computational tractability, we only consider a fixed number of paths at each step, using an estimate
\[\tilde{w}(q')=\max_{q\in E:\delta(q,x_j)\ni q'} \tilde{w}(q'|q)\leq \hat{w}(q')\]
where $E$ is the set of explored states.

The overall procedure for this tokenization schema is as described in Algorithm~\ref{alg:fst}. With the beam search implementation, at most $k\cdot|\mathbf{x}|$ states of the FST will be explored for an input string $\mathbf{x}$. This results in an overall time complexity of $O(k\cdot|\mathbf{x}|\cdot |E|)\cdot T(\mathcal{M}(\mathbf{x}))$, where $T(\mathcal{M})$ is the time complexity for running the language model on $\mathbf{x}$.

\begin{algorithm}[H]
\small
\label{alg:fst}
\caption{Process sequence tokenizer algorithm}
\begin{algorithmic}
\REQUIRE an input string $\mathbf{x}$
\REQUIRE a beam width $k>0$
    \STATE $\mathcal{A} \gets$ an acceptor FST for $\mathbf{x}$
    \IF{$\mathbf{x}$ is not segmented}
        \STATE $\mathcal{T}_{pre} \gets$ a preprocessor to handle punctuation and add word boundaries
        \STATE $\mathcal{T}_{post} \gets$ a postprocessor to convert boundary characters back to spaces
        \STATE $\mathcal{T}_{seg} \gets$ a prebuilt WFST that performs G3-style tonal segmentation
        \STATE $\mathcal{T}_{full} \gets \mathrm{remove\_epsilons}(\mathcal{A} \circ \mathcal{T}_{pre} \circ (\mathcal{T}_{seg})^*\circ \mathcal{T}_{post})$
        \COMMENT{Apply star closure}
        \STATE $\mathcal{M} \gets$ a pretrained neural language model
        \STATE $Q \gets$ a new queue containing only $q_0(\mathcal{T}_{full})$
        \STATE $next \gets \{\}$
        \STATE $\mathbf{\hat{y}} \gets$ dictionary: $q_0\to\epsilon$ \textbf{else} NONE
        \STATE $\hat{w} \gets  \lambda$ dictionary: $q_0\to 0$ \textbf{else} $-\infty$
        \WHILE{$Q$ is not empty}
            \STATE $q \gets \mathrm{pop}(Q)$
            \FOR{$t$ in $\mathrm{outgoing}(q)$}
                \STATE $q' \gets \mathrm{target}(t)$
                \STATE $y' \gets \mathrm{output\_text}(t)$
                \STATE $score\_local \gets \hat{w}(q)+w(t)+\mathcal{M}(y'|\mathbf{x},\mathbf{\hat{y}}[q])$
                \IF{$score\_local > \hat{w}[q']$}
                    \STATE $\hat{w}[q'] \gets score\_local$
                    \STATE $\mathbf{\hat{y}}[q'] \gets \mathbf{\hat{y}}[q]\cdot y'$
                    \STATE $next \gets next \cup \{q'\}$
                \ENDIF
            \ENDFOR
            \IF{$Q$ is empty}
                \STATE \COMMENT{Beam pruning step}
                \FOR{$i$ in $0..k$}
                    \IF{$next$ is empty}
                        \STATE \textbf{continue}
                    \ENDIF
                    \STATE $q_{next} \gets \arg \max_{q\in next} \hat{w}(q)$
                    \STATE $Q.\mathrm{append}(q_{next})$
                    \STATE $next \gets next \setminus \{q_{next}\}$
                \ENDFOR
                \STATE $next \gets \{\}$
            \ENDIF
        \ENDWHILE
        \STATE reweight $\mathcal{T}_{full}$ with the values in $\hat{w}$
        \STATE $\mathcal{T}_{segmented} \gets$ best path through $\mathcal{T}_{full}$
        \STATE
        \COMMENT{$\mathcal{T}_{segmented}$ should output $\mathbf{\hat{y}}[(\arg\max_{q_f\in F} \hat{w}[q_f])]$}
    \ELSE
        \STATE $\mathcal{T}_{segmented} \gets \mathcal{A}$
    \ENDIF
    \STATE $\mathcal{T}_{linearize} \gets$ a prebuilt FST that converts segmentations to the linear format
    \RETURN $\mathrm{output\_text}(\mathrm{best\_path}(\mathcal{T}_{segmented}\circ \mathcal{T}_{linearize}))$
\end{algorithmic}
\end{algorithm}

Further segmenting the lemmatized forms into subword units after separating out the tonal morphology is possible, and running an algorithm like BPE at this stage carries less risk of splitting multiple morphemes across different tokens. This would be similar to Amrhein and Sennrich's experiments using BPE with abstract tokens for non-concatenative morphemes \cite{amrhein2021suitablesubwordsegmentationstrategies}.

\section{Tokenizer Evaluation}

Outside of ASR performance, we will also compare tokenizers on intrinsic metrics: morphological consistency and sparsity. Morphological consistency measures how closely a tokenization scheme reflects a language's morphological structure. Sparsity, on the other hand, is a measure of the information density of a corpus using some tokenizer.

The morphological consistency score, based on the definition from Asgari et al., is defined as the F1-score for the association of tokens with morphemes. The higher the F1 score, the more consistent the tokenizer is with the language's morphology. We compute this pairwise over many samples from the YM dictionary \cite{amithdictionary}, leveraging the dictionary's structure to identify words that share common morphemes.
In particular, we define the following:
\begin{itemize}
    \item A \textbf{True Positive} occurs when a pair of words share at least one common morpheme \textit{and} at least one common token.
    \item A \textbf{False Positive} occurs when a pair of words with \textit{no} morphemes in common share at least one common token.
    \item A \textbf{False Negative} occurs when a pair of words with at least one common morpheme have \textit{no} tokens in common in their representation.
\end{itemize}
From this, we compute the F1 score as \[F_1=\frac{2\cdot TP}{2\cdot TP + FP + FN}\]

In the absence of gold-standard segmentations across the corpus, we consider only verbs in the dictionary, as these have the most robust inflectional information available to us. We adopt the simplifying assumptions that (a) any two inflected forms in the same dictionary entry share a morpheme, (b) any two lexemes with the same inflection (ie. in the same column) share a morpheme, (c) entries with different inflections \textit{and} lexemes share a morpheme if, and only if, they share a common root according to the dictionary or are in columns with overlapping properties\footnote{e.g. the column for \acrshort{neg}+\acrshort{hab} overlaps with \acrshort{hab} since they both share the habitual morpheme.}. Additionally, we ignore enclitics, as these are given separate entries, and would introduce unnecessary complexity to account for in our computations. Since the verb inflections include the primary sources of tonal morphology in YM, we believe these assumptions will give us a reasonable estimate that is directly relevant to our task.

We use a sample size of $n=3300$ pairs of words, taking 650 positive samples from group (a), another 650 positive samples from group (b), and the remaining 2000 samples from group (c). This balance was chosen to have a roughly even split of positive and negative samples in the overall population based on empirical testing of the amount of positive samples from group (c).

Sparsity is an information theoretic measure representing token-level information density. It is defined as a ratio of the entropy of a tokenized corpus to that of the corpus itself. Formally, for tokenizer $\tau$ and corpus $\mathcal{C}$, we define \[\mathrm{Sparsity}(\tau,\mathcal{C})=\frac{H(\tau(\mathcal{C}))}{H(\mathcal{C})}\]
Notably, for our purposes, we assume that the corpus is encoded at a character level, so a character-level tokenizer should have a sparsity of exactly 1, modulo padding or other special tokens. This gives us a more suitable baseline for the subword tokenization schemes we're considering, as opposed to considering word-level entropy as a baseline.
Tokenizers with a higher sparsity tend to have more tokens that occur infrequently within the corpus, and as such a language model will have limited opportunity to learn a meaningful embedding for those tokens.

We hypothesize that our novel tokenizers are not simply better compression methods, but that they are good representations of YM morphology. Quantifying both information density and morphological consistency would thus provide evidence to support the standing hypothesis that tokenizers that are well-aligned with morphology produce models that perform better in downstream tasks. These metrics will allow us to quantify both of these dimensions.

\section{Dataset}
The dataset used in this project is a corpus of spoken Yolox\'ochitl Mixtec compiled by Jonathan Amith and Rey Castillo Garc\'ia. At the time of writing, the full corpus has not been publicly released, and was used with Amith's permission as part of a project to expand the corpus with a full segmentation tier. Due to limited data availability, the ASR models built for this task will be trained on the unsegmented tier, as this will allow us to get comparable results to the original ASR paper with a sizable split of the data.

The corpus we will use consists of 52.4 hours of transcribed YM speech, with 42 hours in the training split and 5.2 hours each in the evaluation split and the test split. The text is written in the practical orthography, which reflects underlying morphology, but not phonological changes. The data was cleaned automatically with a preprocessing script we integrated into the ESPnet recipe. In keeping with the goals for the segmentation corpus, our data preserves all original punctuation from the original transcription. However, elided tones--usually marked with parenthesis in the unsegmented practical orthography--are unmarked, reflecting the segmentation. Additionally, content in square brackets, which marks revisions added by the transcribers for clarity that was not in the original audio, was removed to maintain consistency with the ASR task, as was done in Shi's original paper.

\section{Experiments}

All configurations use 12 encoder blocks and 6 decoder blocks, with a 2048 dimension feed-forward layer and a dropout rate of 0.1. The input layer is a convolutional 2D subsampling layer. We use a hybrid CTC/Attention architecture with a CTC ratio of 0.3. The loss function is cross-entropy loss with label smoothing.
Training is performed with an Adam optimizer with a learning rate of 1.0 and a Noam scheduler with 25000 warm-up steps, up to a max epoch of 100. The best model is selected on the basis of performance on the evaluation set.
The input acoustic features are 83-dimensional log-Mel filterbanks with pitch features.
The transformer models all have a four-head multi-attention layer with 256 dimensions in the encoder. The conformer models all have encoders with an eight-head multi-attention layer with 512 dimensions and a CNN module with a kernel size of 15.
Models that utilize SpecAugmentation use bicubic time warping with a window size of 5 frames, two frequency masks with a width range of $[0,30]$, and two time masks with a width range of $[0,40]$.

Our first round of experiments will be to determine the best configuration for model training, particularly comparing transformer and conformer architecture, as well as varying the subsampling factor and the initialization method. We will test subsampling factors of 2x, 4x, and 6x for the convolutional subsampling layer, and models will be run with both Chainer initialization and Xavier Uniform initialization. We will also run the transformer models both with and without SpecAugment. All of these experiments will be run using a Unigram tokenizer with a vocab size of 500.

Following this, we will use the best-performing configuration to train models for each of our tokenizers, as well as baselines with word-level tokenization, character-level tokenization, and three linear subword tokenizers---BPE, Unigram, and WordPiece---which will be tested for vocabulary sizes of 500 and 1000. The three top-performing linear tokenizers will also be combined with with Segment-Melody and Process Sequence tokenizers to produce augmented tokenization schemes, which will also be evaluated. CER and WER will be computed five times for each tokenizer to perform statistical tests for significance.

\chapter{Results}

\section{Model Comparison}

\begin{table}[b]

\centering
\begin{tabular}{l|c|c|c|cc}
\textbf{Family} & \textbf{Initialization} & \textbf{SpecAug?} & \textbf{Subsampling} & \textbf{CER} & \textbf{WER} \\ \hline
Transformer  & Chainer   & No  & 4x & 14.2         & 32.5          \\
             & Chainer   & Yes & 4x & 10.2         & 24.1          \\
             & Xavier    & No  & 4x & 11.8         & 27.9          \\
             & Xavier    & Yes & 2x & 10.9         & 25.0          \\
             & Xavier    & Yes & 4x & \textbf{9.2} & \textbf{22.2} \\
             & Xavier    & Yes & 6x & 10.2         & 23.6          \\ \hline
Conformer    & Xavier    & Yes & 4x & 89.6         & 91.8    
\end{tabular}
\caption{Comparison of model performance using a Unigram tokenizer with vocabulary size 500.}
\end{table}

Our experiments indicate that a transformer model with Xavier Uniform initialization and a subsampling factor of 4 in the input layer performs best. Controlling for other variables, Xavier Uniform initialization consistently shows better performance over Chainer initialization, and the use of SpecAugment also shows an improvement in performance regardless of the initialization method. Subsampling factors of 2 and 6 seemed to decrease performance, suggesting that a factor of 4 is optimal for this task.

The conformer models performed considerably worse than the transformer models under identical conditions, yielding a CER of almost 90\%. According to Shi, the conformer models generally require multiple rounds of fine-tuning to achieve optimal performance. Consequently, due to the time constraints of this project and the negligible room for improvement over the transformer model, we chose to use the transformer model with Xavier Uniform initialization and SpecAugment for the remaining experiments.

\section{Tokenizer Ratings}
\label{exp:metrics}

\begin{table}[]
    \centering
    \begin{tabular}{r|l|c|c}
    \toprule
    \textbf{Family} & \textbf{Tokenizer} & {\textbf{Morph.-F1} ($\uparrow$)} & {\textbf{Sparsity} ($\downarrow$)} \\
    \midrule
(Simple)        & \textit{Word}       & \textit{0.249} & \textit{2.252} \\
                & \textit{Character}  & \textit{0.635} & \textit{1.000} \\
\hline
SentencePiece   & BPE/500             & 0.673 & 1.411 \\
                & BPE/1k              & 0.673 & 1.409 \\
                & Unigram/500         & 0.665 & 1.351 \\
                & Unigram/1k          & 0.665 & 1.351 \\
\hline
WordPiece (HF)  & WordPiece/500       & 0.662 & 1.741 \\
                & WordPiece/1k        & 0.644 & 1.762 \\
\hline
Segment-Melody  & SegMel Base         & 0.716 & 1.623 \\
                & SegMel+BPE/500      & 0.642 & 1.460 \\
                & SegMel+BPE/1k       & 0.642 & 1.457 \\
                & SegMel+Unigram/500  & 0.636 & 1.432 \\
                & SegMel+Unigram/1k   & 0.637 & 1.432 \\
\hline
Process Sequence& ProcSeq Base        & 0.648 & 1.419 \\
                & ProcSeq+BPE/500     & 0.640 & 1.317 \\
                & ProcSeq+BPE/1k      & 0.640 & 1.315 \\
                & ProcSeq+Unigram/500 & 0.636 & 1.273 \\
                & ProcSeq+Unigram/1k  & 0.636 & 1.273 \\
    \bottomrule
    \end{tabular}
    \caption{Comparison of intrinsic metrics across the tokenizers used in our experiments. Word and Character tokenizers are included as control points: word-level tokenization should have low F1 scores and high sparsity, while character-level tokenization should have somewhat high F1 and a sparsity of exactly 1.}
    \label{tab:table-metrics}
\end{table}

Of the tokenizers we examined, we found a median Morph.-F1 score of 0.642, and a median sparsity of 1.415. All four SentencePiece tokenizers we used had both better (higher) Morph.-F1 scores than the median and better (lower) sparsity than the median, making them balanced choices across both metrics. Meanwhile, the WordPiece and Segment-Melody tokenizers consistently had high Morph.-F1 scores with sparsity above the median.
The base tokenizers for both SegMel and ProcSeq have Morph.-F1 scores above the median, indicating that they are fairly consistent with the morphology, which is as desired.

Excluding the baseline, the ProcSeq+Unigram tokenizers showed the lowest sparsity, and the WordPiece tokenizers showed the highest sparsity. For Morphological F1 scores, the base SegMel tokenizer exhibited the highest score, while the ProcSeq+Unigram tokenizers exhibited the lowest scores.

The full results are displayed in Table~\ref{tab:table-metrics}.

\section{Tokenizer Performance in ASR}

\begin{table}[t]

\centering
\begin{tabular}{r|l|c|cc}
\toprule
\textbf{Family} & \textbf{Tokenizer} & \textbf{Vocab Size} & {\textbf{CER}} & {\textbf{WER}} \\ 
\midrule
(Simple)        & Word               & 23331   & {19.5}       & {34.6}         \\
                & Character          & 53      & {11.8}       & {29.6}         \\
\hline
SentencePiece   & BPE/500            & 500     & \textbf{8.5} & {22.9}         \\
                & BPE/1k             & 1000    & {10.5}        & {26.2}         \\
                & Unigram/500        & 500     & {8.6}        & {23.0}         \\
                & Unigram/1k         & 1000    & {8.7}        & {23.0}         \\
\hline
WordPiece (HF)  & WordPiece/500      & 500     & {16.4}        & {95.6}         \\
                & WordPiece/1k       & 1000    & {16.7}        & {94.4}         \\
\hline
Segment-Melody  & SegMel Base    & 3195    & {8.8}        & \textbf{22.5}  \\
                & SegMel+BPE/500     & 500     & {8.8}        & {23.4}         \\
                & SegMel+BPE/1k      & 1000    & {11.0}       & {26.9}         \\
                & SegMel+Unigram/500 & 500     & {9.0}        & {23.7}         \\
                & SegMel+Unigram/1k  & 1000    & {9.1}        & {23.8}         \\
\hline
Process Sequence& ProcSeq Base   & 5759    & {9.6}        & {24.6}           \\
                & ProcSeq+BPE/500    & 500     & {14.9}       & {34.4}           \\
                & ProcSeq+BPE/1k     & 1000    & {15.3}       & {34.8}           \\
                & ProcSeq+Unigram/500& 500     & {13.9}       & {33.1}           \\
                & ProcSeq+Unigram/1k & 1000    & {14.2}       & {33.0}           \\
\bottomrule
\end{tabular}
\caption{Mean CER and WER on the test split for models trained on different tokenizers, averaged over five runs.}
\label{tab:asr-results}
\end{table}

Our results for the ASR task indicate that using BPE with a vocab size of 500 produces the lowest character error rate, and that the Segment-and-Melody tokenizer produces the lowest word error rate. The four SentencePiece tokenizers (BPE and Unigram), the novel Segment-and-Melody and Process Sequence tokenizers, and the augmented SegMel+SP tokenizers are all within the 10\% CER limit set by Amith et al. to achieve an optimal balance between automated transcription and human proofreading \cite{amith2021end}.

We use the Wilcoxon ranked-sign test with $n=5$ to evaluate the significance of our results. Assuming all experiments to be independent, we consider each pair of tokenizers to have some distribution of $X_i-Y_i$, where $X_i$ is the first tokenizer's performance in experiment $i$, and $Y_i$ is the second tokenizer's performance in experiment $Y_i$.
The null hypothesis for this test is $\mathbb{E}[X_i-Y_i]=0$, indicating that there is no expected difference in the performance distributions between the two tokenizers, which we reject when we get a one-tailed $p$ value below $\alpha=0.05$.

Comparing the BPE/500 tokenizer with the base Segment-Melody tokenizer, we find that the BPE models are significantly better in terms of character error rate ($T^+=15, p=0.031$) than the SegMel models, while the SegMel models are significantly better in terms of word error rate ($T^+=0, p=0.031$) than the BPE models. The BPE model with a vocabulary size of 500 performed significantly better than the model with a vocabulary size of 1000 in both CER and WER (both $T^+=0,p=0.031$), while the difference between the two Unigram models is not found to be significant in terms of either CER ($T^+=10,p=0.313$) or WER ($T^+=12,p=0.156$). Comparing BPE and Unigram models with the same vocabulary size, the difference in results between BPE/500 and Unigram/500 is also not found to be significant for either CER ($T^+=10,p=0.313$) or WER ($T^+=12,p=0.156$). Similar results do not indicate significant differences in the performance of Unigram-augmented models with different vocab sizes, with $T^+=12,p=0.156$ for both CER and WER on SegMel+Unigram and ProcSeq+Unigram.

The WordPiece tokenizers performed significantly worse in CER for all tokenizers except word-level, where it performed significantly better (vs Word: $T^+=0, p=0.031$, all others: $T^+=15, p=0.031$). Notably, the WordPiece tokenizers exhibit a remarkably high WER, approaching 100\% as opposed to the other tokenizers' WER scores of under 35\%.

Augmenting the novel tokenizers with Unigram resulted in a significant increase in both CER and WER (all $T^+=15,p=0.031$). Neither novel tokenizer augmented with Unigram showed a significant difference in performance between different vocab sizes (WER and CER: $T^+=3,p=0.156$).
Interestingly, augmenting the tokenizers with BPE led to divergent results. For the SegMel tokenizer, we see a significant increase in WER ($T^+=15, p=0.031$) for the BPE-augmented version with vocab size 500, but a \textit{decrease} in CER that does not meet the threshold for significance ($T^+=6, p=0.406$). The SegMel+BPE/500 model performed significantly better than the SegMel+Unigram models in CER and WER (all $T^+=0, p=0.031$). Increasing the BPE vocab size to 1000 resulted in a significant underperformance in both WER and CER compared to all other augmented Segmel models, including Segmel+BPE/500 (all $T^+=0, p=0.031$).
On the other hand, for the ProcSeq tokenizer, we see significant increases in both CER and WER (both $T^+=15,p=0.031$) by adding a BPE step of any vocab size. Additionally, the ProcSeq+BPE models performed \textit{worse} than the ProcSeq+Unigram models in CER and WER, with the difference in CER reaching a degree of significance (all $T^+=15,p=0.031$). The difference in WER between the BPE/500 and Unigram/1000, as well as between the BPE/1000 and Unigram/500 augmented ProcSeq tokenizers fell short of significance with $T^+=14, p=0.063$. The change in vocab size did not exhibit a significant difference in performance for ProcSeq+BPE tokenizers (WER: $T^+=5,p=0.313$, CER: $T^+=4,p=0.219$).
All novel tokenizers show a significant improvement over word-level tokenization, and all novel tokenizers, except the augmented Process Sequence tokenizers, also show a significant improvement over character-level tokenization.

\vspace*{-12pt}
\subsection{Correlation between ASR performance and intrinsic metrics}
\vspace*{-4pt}

\begin{figure}
    \centering
    \includegraphics[width=0.9\linewidth]{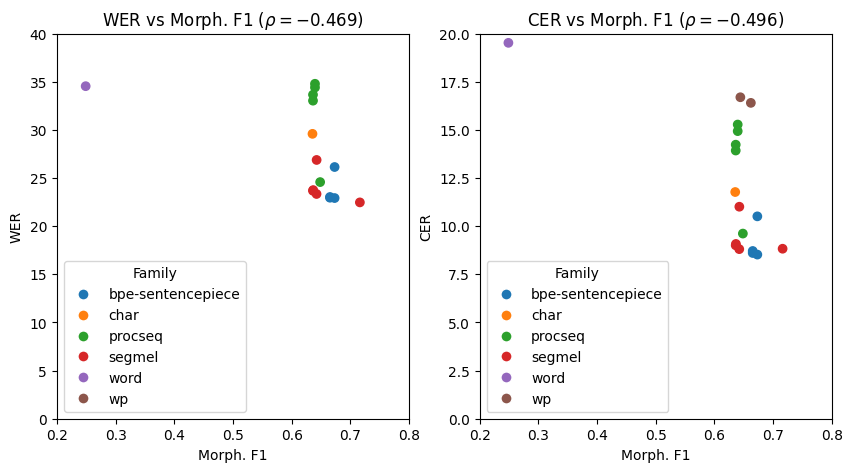}
    \includegraphics[width=0.9\linewidth]{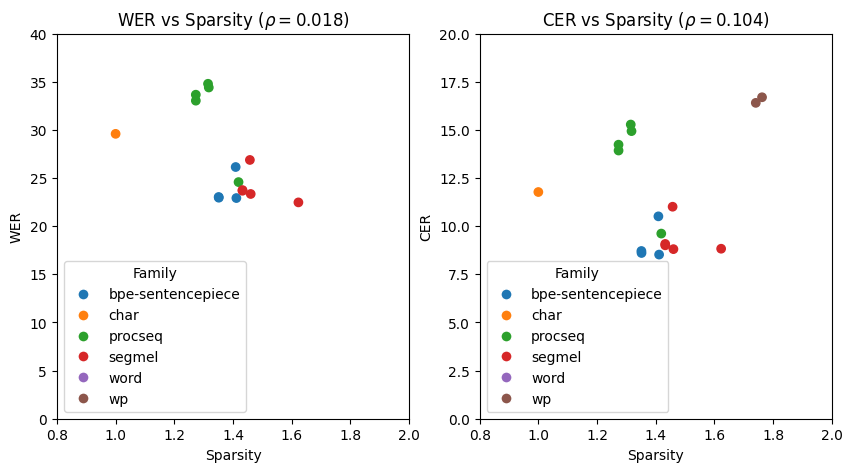}
    \caption{Comparison of ASR performance metrics WER (left) and CER (right) and intrinsic tokenizer metrics Morph.-F1 (top) and Sparsity (bottom). Each point on the plot represents a tokenizer, using the average WER and CER from Table~\ref{tab:asr-results}. Note that outliers may not appear in the graph due to cropping.}
    \label{fig:correlation_morph_f1}
\end{figure}

Comparing the ASR performance against the intrinsic metrics computed in section~\ref{exp:metrics}, we find that the Morphological F1 score is loosely correlated with both WER and CER in the ASR task. Using Spearman's rank correlation coefficient, we get $\rho=-0.469$ for Morph.-F1 vs. WER ($p=0.049$) and $\rho=-0.496$ for Morph.-F1 vs. CER ($p=0.036$), both of which are significant. Computing the same statistic for Sparsity, we find no significant correlation, with $\rho=0.018$ for Sparsity vs. WER ($p=0.945$) and $\rho=0.104$ for Sparsity vs. CER ($p=0.681$). 
Scatter plots illustrating these comparisons are presented in Figure~\ref{fig:correlation_morph_f1}.

\chapter{Discussion}

The observed correlation between error rate and Morphological F1 scores supports our hypothesis that tokenizers which represent morphology produce more accurate ASR models. In particular, the base Segment-Melody tokenizer outperforming the best traditional subword tokenizer in terms of ASR WER suggests that tokenizers designed specifically for a language's morphology can produce more accurate ASR results. The discrepancy between WER and CER for the SegMel tokenizer compared to the BPE tokenizer, which scored best in CER and second-best in WER, suggests the possibility that small, systemic errors are more common with BPE, while the errors produced with SegMel have more incorrect characters per incorrect word. This means that errors are more tightly clustered: where SegMel predicts one incorrect word with several incorrect characters, BPE might predict several words incorrectly with only one character off. For example, BPE might substitute \orth{\olang{kwi3i3}} for \orth{\olang{kwe3e3}} consistently across many occurrences, whereas SegMel might mistake \orth{\olang{kwi3i3}} for \orth{\olang{se'1e3}} across a handful of instances. While Amith et al. use CER as the primary metric for analyzing the effectiveness of an ASR model for corpus transcription \cite{amith2021end}, WER is arguably more useful to a human annotator because revising individual incorrect words requires less time than combing through the transcription to fix uniform errors. CER is still an important and useful metric for ASR because it is more granular and reflects partially correct words, but between ASR models with comparably low CER, the one with lower WER is better suited to the corpus annotation task.

\section{Linear Tokenizer Evaluation}

WordPiece exhibited the worst performance out of the subword tokenizers, although this seems to be largely a result of the preprocessing and postprocessing schemes naively inserting spaces around hypen (\texttt{-}) and equals (\texttt{=}) characters. Manually removing these extra spaces reduced the error rates to levels comparable to the SentencePiece tokenizers (around 9.6\% CER and 28\% WER on the dev split), albeit still slightly higher than the best-scoring tokenizers.
Initially, I implemented a basic training script using the HuggingFace implementation of BERT WordPiece for compatibility with ESPnet. I used a default normalizer class for BERT, which likely made language-specific assumptions about how different punctuation symbols interact with text, such as always splitting apostrophes (\texttt{'}) from their context as a separate token and always padding hyphens and equals signs with spaces. These normalization assumptions did not appear in the BPE and Unigram models, which were constructed using the SentencePiece command-line utility \texttt{spm\_train} without under-the-hood preprocessing or normalization.

Of the linear tokenizers, BPE with a vocab size of 500 performs the best by a slim margin, followed by the two Unigram models, whereas BPE with a vocab size of 1000 lags significantly behind. This indicates that the tokenizations generated by the Unigram algorithm are more stable with different vocab sizes than those generated by BPE. The splits that the Unigram algorithm has made to reach a vocabulary size of 1000 have already optimized the per-token information density, so reducing the vocabulary size to 500 does not produce a significant increase in information density. On the other hand, the BPE algorithm, in performing a greedy search from the bottom up, is likely encountering an increased possibility of merging adjacent tokens that are together by chance rather than because of a pattern. As such, while the first set of iterations to reach a vocab size of 500 keeps the tokens shorter and more informative, the next set of iterations begins to merge tokens in ways that lose connections between related words. This issue has been observed and documented before, and is a major reason behind BPE being suboptimal in some NLP tasks.

\section{Novel Tokenizer Evaluation}

The Process Sequence tokenizer performed the worst out of the tokenizers tested that did not exceed a character error rate of 10\%. Since we did not have the dataset to fully evaluate the segmentation model, it is possible that the segmentations produced by our tokenizer were of low quality and introduced noise into the model, producing a cascade of errors. The low Morph.-F1 scores for the base ProcSeq tokenizer, beating the median but not beating the BPE and Unigram scores, support this theory, as it suggests that the segmentation model may not have correctly learned the morphological connections between related words.
Another potential source of error is the number of possible token sequences that could represent the same melody: each mora gets a separate token for its tone, meaning bimoraic words are represented as three tokens and trimoraic words are represented as four. Additionally, multiple tokens are possible for the same surface melody, which may produce less certainty if, e.g., the model is choosing between \texttt{1>4}, \texttt{3>4}, and \texttt{4>4} for a tone \orth{4} in the audio. As a result, the compounding uncertainties may result in low confidence predictions of the tone melody, and in turn erroneous predictions of the surface forms. The point about token disambiguation may also be evaluated in more depth with a segmentation dataset.

The augmented ProcSeq+Unigram tokenizers showed low sparsity and low Morph.-F1 scores, which indicates that while they are effective at compression, they do not closely represent morphology. While morphology was the original goal, this is an interesting result, and may be worth further study of the information theoretical impact of nonlinear tokenizers to encode languages with non-concatenative morphology.

Another notable result is that the Process Sequence tokenizer took a very long time to run, even after implementing a cache for LLM predictions over a common prefix. Preprocessing all 24k lines of the train set took roughly nine hours, while the SegMel tokenizer only took about 2.5 minutes on the same split. We suspect that this is at least in part a result of the FSTs constructed for our runtime not being fully optimized and determinized: we found cases where a single input-output pair had multiple arcs between the same initial and final states. We had attempted to troubleshoot this issue, but due to time constraints, we were unable to resolve it completely.

A somewhat surprising result is that running BPE on top of either of our novel nonlinear tokenizers resulted in worse performance than running the novel tokenizers or BPE independently. This goes against the results of Asgari et al., 2025, and Amrhein and Sennrich, 2021, which indicated better performance for the BPE-augmented tokenizers over standard BPE \cite{asgari2025morphbpemorphoawaretokenizerbridging,amrhein2021suitablesubwordsegmentationstrategies}. Further research could be done into whether this is a result of how tonal morphology patterns in languages like YM in comparison to how the non-concatenative morphology patterns in their experiments with Arabic, or whether this is a result of the distribution of which YM words exhibit tonal morphology. Compared to a language like Arabic, YM has fewer combinations for derivation and inflection on the same root, so it's possible that BPE alone does a sufficient job at capturing this information in YM where it falls behind with Arabic. Further study would need to be done to compare the efficacy of unaugmented tokenizers such as SegMel in other languages with more abundant non-concatenative morphology.

The lack of a significant correlation between sparsity and ASR error rate suggests that sparsity is not an effective predictor of ASR performance. Morphological consistency F1 scores are a better predictor, albeit still not strictly monotonic: tokenizers with higher Morph.-F1 scores can generally be expected to perform better in ASR, although this is not strictly guaranteed. We recommend analysis of Morph.-F1 scores in future tokenization research for morphologically rich languages, including languages with non-concatenative morphology.

Frustratingly, we were unable to complete a gold-standard segmentation dataset for YM within the time constraints for my project. This severely limited our ability to evaluate the performance of the models we built, including the inability to produce any baselines for the segmentation task. Hopefully, the dataset will be completed in short order by my colleagues on the wav2gloss team, as using the Process Sequence tokenizer to produce an end-to-end speech-to-segmentation model from existing ASR infrastructure could be a net improvement, even if the error rates on the unsegmented ASR task remain higher than other tokenization methods.

\chapter{Conclusion}

Our results suggest that nonlinear tokenizers designed around a language's morphology can lower ASR word error rates compared to conventional subword tokenizers. The character error rates for ASR models trained using these tokenizers tend to remain within the 10\% margin suggested by Amith et al. for an optimal workflow, which remains competitive with ASR models trained on Unigram or BPE tokens.

We were surprised to find that adding a BPE or Unigram step to our novel tokenizers reduces the performance in the ASR task to a signification degree ($\alpha=0.05$) compared to the original BPE and Unigram runs. The results of Amrhein and Sennrich, 2021, suggested that BPE over an abstract representation would improve performance \cite{amrhein2021suitablesubwordsegmentationstrategies}. However, it is possible that the difference in performance stems from a difference in the datasets: our training dataset consists of ~23k segments of low-resource language data in YM, while Amrhein and Sennrich used a corpus of ~4.6M segments of parallel data between two high-resource languages. Furthermore, since the abstract forms they utilized contained a text description of the morphemes, it is possible that the BPE model was better able to capture the morphology for Amrhein and Sennrich in fewer tokens, while the BPE models run on top of our novel tokenizers for YM may have split meaningful morphemes less consistently.

We discovered that Morphological consistency F1 scores, based on the formulation by Asgari et al. \cite{asgari2025morphbpemorphoawaretokenizerbridging}, correlate with ASR performance to a signification degree. This supports prior findings that tokenization schemes that correspond well with morphology produce models that perform better on downstream tasks \cite{toraman2023impact,bostrom2020byte}, and extend this quantifiably to the ASR task. We recommend using Morphological F1 scores as a measure of morphological consistency in future tokenizer research and also as a rough predictor of downstream performance.
Sparsity, on the other hand, was not found to correlate with ASR performance in our experiments, which suggests that information density is not as important of a factor in tasks like ASR compared to morphology.

We also recommend weighting WER over CER in evaluation of tasks involving human annotators revising machine-generated annotations. While CER is a useful metric to identify the severity of errors the machine makes, the annotator will be making revisions on a larger scale, considering whole words in a larger utterance, which makes WER a somewhat more useful metric for this task. An annotator would likely spend less time correcting a single word with a high edit distance from the intended word than they would spend correcting several erroneous words with low edit distances. We posit that, given multiple ASR systems of a sufficiently low CER (eg. under 10\%, per \cite{amith2021end}), the one with the lowest WER is optimal for the machine-assisted annotation task, not necessarily the one with the lowest CER. Further research into this hypothesis is necessary to study the full impact of CER and WER in an annotation task.

\section{Future Research}

Unfortunately, we were unable to evaluate the performance of the ASR models we trained in segmentation tasks. Based on the findings of Toraman et al., we predict that the segmentation quality of the model we used for the ProcSeq tokenizer has room for improvement, as the tokenizer's underperformance is likely a result of poor consistency with the underlying morphology \cite{toraman2023impact}. We hypothesize that with improved fine-tuning on the segmentation model, the ProcSeq tokenizer will be able to improve performance on the ASR task and to produce accurate segmentations from speech.
Hopefully, future work will be able to evaluate a baseline and make improvements as necessary.

Having a gold-standard segmentation dataset for YM will be essential for full research and analysis of these novel tokenization methods. We present a proof-of-concept that nonlinear tokenization methods are practical for working with languages exhibiting non-concatenative morphology, which we hope can be leveraged in future research once gold-standard segmentation data becomes available. The utility of these kinds of tokenizers lies in their potential to have higher accuracy in producing G3-style segmentations directly from spoken audio; this potential can only be evaluated quantitatively once segmentation data is available.

A set of 1300 sentences from the Amith's YM corpus has been selected to provide gold-standard segmentation and gloss data for evaluation purposes. At present, these examples have been identified and key vocabulary items have been disambiguated, but production of a full segmentation and gloss for all selected sentences remains to be completed.
Once this dataset is available, further research can be conducted on the effectiveness of these tokenizers in building ASR models that can learn segmentations. In particular, investigating whether or not the Process Sequence tokenizer, after being trained on unsegmented data, can produce usable segmentations will be insightful into the requirements for constructing a direct wave-to-segmentation pipeline. Once a direct wave-to-segmentation pipeline is available, it can be evaluated and compared against a cascaded system that feeds ASR output into a text-to-text segmentation model, and ultimately determine which architecture is better suited to the wav2gloss task.

Further optimizations of the FST pipeline could be considered, both in terms of reducing the number of states and arcs to remove redundancy as well as in terms of reweighting the FST. We had considered a procedure to train the FST weights based on the licensed segmentations of the words in Amith's dictionary, but due to the redundant arcs, we did not see much improvement. We believe that given more time, further optimization and reweighting is possible and can improve both runtime and accuracy.

While the tokenizers we present are specific to YM, similar methods can be used to construct nonlinear tokenizers for other languages. FSTs and regular expressions, both commonly used and well-documented frameworks, form the underlying backbone of our novel tokenizers. An FST-based pipeline has, in effect, the same expressive power as regular expressions, and therefore any regular patterns can be pulled from the interior of words and added as a token to the end of the sequence, with the added benefit of being fully differentiable, which is useful when working with ML pipelines. Our tokenizers utilized these methods to produce abstract tokenization schemes for training ML models for language processing, which do not map strictly linearly onto the original string. The results of our research suggest that such nonlinear tokenizers can produce stronger, more accurate language models for languages exhibiting non-concatenative morphology.

\appendix

\printglossary[type=\acronymtype,title={\chapter{Glossary of Linguistic Abbreviations}\vspace*{-2cm}}]

\chapter{Code info}
The code for this project is available on GitHub:

\begin{itemize}
    \item ESPnet support for tokenizers: 
\texttt{\href{https://github.com/Chris-dash-T4/espnet/tree/wav2gloss}{https://github.com/Chris-dash-T4/espnet}} (branch \texttt{wav2gloss})
    \item Data cleanup and processing: \texttt{\href{https://github.com/dmort27/mixtec-segment-gloss-disambiguation}{https://github.com/dmort27/mixtec-segment
    -gloss-disambiguation}}
    \item FST backend implementation: \texttt{\href{https://github.com/Chris-dash-T4/mixtec-fst-epitran.rs}{https://github.com/Chris-dash-T4/mixtec
    -fst-epitran.rs}}
\end{itemize}

\backmatter

\printbibliography[title={Bibliography
\addcontentsline{toc}{chapter}{Bibliography}
}]

\end{document}